\def\cvprPaperID{5988}
\def\confName{CVPR}
\def\confYear{2023}

\def\paperTitle{Learning to Retain while Acquiring: Combating Distribution-Shift in\\Adversarial Data-Free Knowledge Distillation}

\def\authorBlock{
    Gaurav Patel\textsuperscript{$\dagger$}\quad
    Konda Reddy Mopuri\textsuperscript{$\ddagger$}\quad
    Qiang Qiu\textsuperscript{$\dagger$}\\
    \textsuperscript{$\dagger$}Purdue University\quad\textsuperscript{$\ddagger$}Indian Institute of Technology Hyderabad\\
    {\tt\small \{gpatel10, qqiu\}@purdue.edu, krmopuri@ai.iith.ac.in} 
}

\newif\ifreview 
\newif\ifarxiv \newcommand{\arxiv}{\arxivtrue}
\newif\ifcamera 
\newif\ifrebuttal 

\arxiv

\pdfoutput=1
\documentclass[10pt,twocolumn,letterpaper]{article}
\ifreview \usepackage[review]{cvpr} \fi
\ifarxiv \usepackage[pagenumbers]{cvpr} \fi
\ifrebuttal \usepackage[rebuttal]{cvpr} \fi
\ifcamera \usepackage{cvpr} \fi

\usepackage{graphicx}
\usepackage{amsmath}
\usepackage{amssymb}
\usepackage{booktabs}

\usepackage{times}
\usepackage{microtype}
\usepackage{epsfig}
\usepackage[table,xcdraw]{xcolor}
\usepackage{caption}
\usepackage{float}
\usepackage{placeins}
\usepackage{color, colortbl}
\usepackage{stfloats}
\usepackage{enumitem}
\usepackage{tabularx}
\usepackage{xstring}
\usepackage{multirow}
\usepackage{xspace}
\usepackage{url}
\usepackage{subcaption}
\usepackage{xcolor}
\usepackage[hang,flushmargin]{footmisc}
\usepackage[english]{babel}
\usepackage{amsthm}
\usepackage{amssymb}
\usepackage{booktabs}
\usepackage{graphicx}
\usepackage[normalem]{ulem}
\usepackage{sidecap}

\useunder{\uline}{\ul}{}

\usepackage{algorithm2e}
\RestyleAlgo{ruled}
\usepackage{amsmath}
\DeclareMathOperator*{\argmax}{arg\,max}

\SetKwComment{Comment}{/* }{ */}

\DeclareMathOperator{\softmax}{Softmax}

\ifcamera \usepackage[accsupp]{axessibility} \fi

% ...

\newtheorem{innercustomthm}{Theorem}
\newenvironment{customthm}[1]
  {\renewcommand\theinnercustomthm{#1}\innercustomthm}
  {\endinnercustomthm}

\newtheorem{innercustomlemma}{Lemma}
\newenvironment{customlemma}[1]
  {\renewcommand\theinnercustomlemma{#1}\innercustomlemma}
  {\endinnercustomlemma}

\ifarxiv  \fi

\newcommand{\R}[1]{{%
    \textbf{%
        \ifstrequal{#1}{1}{\textcolor{red}{R#1}}{%
        \ifstrequal{#1}{2}{\textcolor{blue}{R#1}}{%
        \ifstrequal{#1}{3}{\textcolor{magenta}{R#1}}{%
        \ifstrequal{#1}{4}{\textcolor{teal}{R#1}}{%
                           \textcolor{cyan}{R#1}%
        }}}}%
    }%
}}

\usepackage{xr-hyper}

\makeatletter
\newcommand*{\addFileDependency}[1]{
  \typeout{(#1)}
  \@addtofilelist{#1}
  \IfFileExists{#1}{}{\typeout{No file #1.}}
}

\makeatother

\usepackage[pagebackref,breaklinks,colorlinks]{hyperref}
\usepackage[capitalize]{cleveref}
\crefname{section}{Sec.}{Secs.}
\crefname{table}{Table}{Tables}
\crefname{figure}{Fig.}{Figs.}

\begin{document}
\title{\paperTitle}
\author{\authorBlock}
\maketitle

\begin{abstract}
Data-free Knowledge Distillation (DFKD) has gained popularity recently, with the fundamental idea of carrying out knowledge transfer from a Teacher neural network to a Student neural network in the absence of training data. However, in the Adversarial DFKD framework, the student network’s accuracy, suffers due to the non-stationary distribution of the pseudo-samples under multiple generator updates. To this end, at every generator update, we aim to maintain the student’s performance on previously encountered examples while acquiring knowledge from samples of the current distribution. Thus, we propose a meta-learning inspired framework by treating the task of Knowledge-Acquisition (learning from newly generated samples) and Knowledge-Retention (retaining knowledge on previously met samples) as meta-train and meta-test, respectively. Hence, we dub our method as Learning to Retain while Acquiring. Moreover, we identify an implicit aligning factor between the Knowledge-Retention and Knowledge-Acquisition tasks indicating that the proposed student update strategy enforces a common gradient direction for both tasks, alleviating interference between the two objectives. Finally, we support our hypothesis by exhibiting extensive evaluation and comparison of our method with prior arts on multiple datasets.
\end{abstract}
\section{Introduction}
\label{sec:intro}
\begin{figure}[t]
    \centering
    \includegraphics[width=\columnwidth]{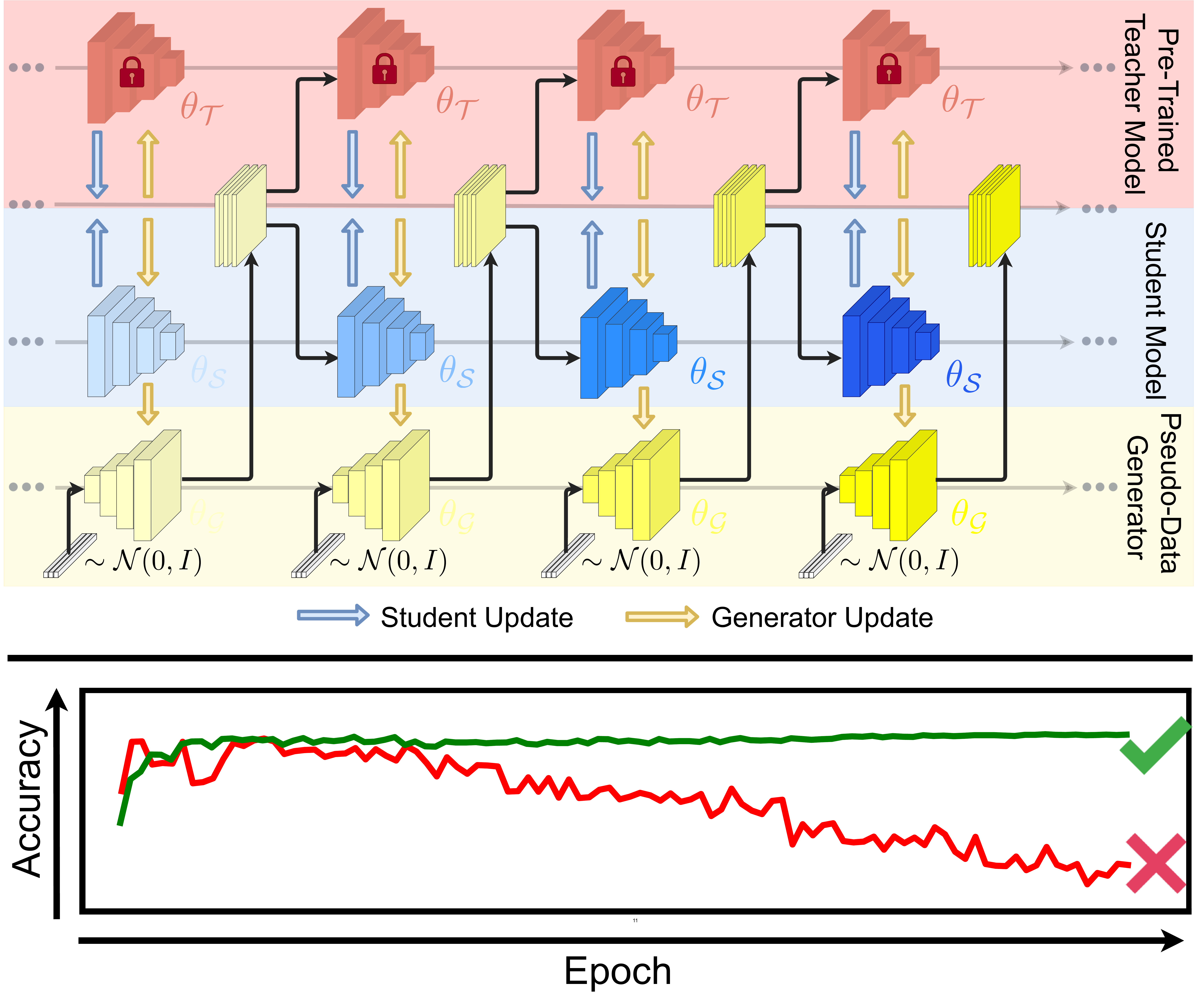}
    \caption{The \emph{top-section} represents the learning evolution of a generic Adversarial Data-Free Knowledge Distillation framework; the color-intensity variation signifies the change in the distribution of the pseudo-samples, the student network, and the generator network over the learning epochs. Under the variation in the distribution of the pseudo-samples, the \emph{bottom-section} shows the learning curves for cases when the student accuracy degrades (shown in {\color[HTML]{ff0000}Red}), which is undesirable, and when the student accuracy is maintained, if not improved, as proposed (shown in {\color[HTML]{008000}Green}).}
    \label{fig:intro}
\end{figure}

The primary goal of Data-Free Knowledge Distillation (DFKD) is to acquire a trustworthy alternative dataset for knowledge distillation. The quality of knowledge transfer depends on the caliber of these alternative samples. Noise optimization~\cite{nayak2019zero,yin2020dreaming,fang2021contrastive} and generative reconstruction~\cite{DAFL,micaelli2019zero,fang2019data} are the two primary ways to replace the original training data used in the distillation process with synthetic or pseudo samples. Adversarial DFKD methods~\cite{fang2019data,liu2021zero,micaelli2019zero} investigate an adversarial exploration framework to seek pseudo-samples. In such a paradigm, the Generator ($\mathcal{G}$) is used to create pseudo-samples as surrogates to perform knowledge distillation/transfer, and the Teacher-Student ($\mathcal{T}$-$\mathcal{S}$) setup acts as a joint discriminator to penalize and update generator parameters ($\mathcal{\theta_{G}}$) (Figure \ref{fig:intro}).
In the adversarial framework, the generator explores the input space to find suitable pseudo-samples as the distillation progresses. Consequently, the distribution of the generated samples consistently keeps changing during the process due to the generator updates \cite{seff2017continual}. From the student network's perspective, at each iteration, the pseudo samples seem to be generated from different generator parameters ($\mathcal{\theta_{G}}$). Hence, the convergence of the student network gets hampered due to successive distributional alterations over time~\cite{thanh2020catastrophic}, as depicted by the red curve in Figure~\ref{fig:intro}. This observation hints that updating the student network, solely using the samples generated from the current generator parameters is not adequate to generalize the student. Moreover, the student forgets the knowledge acquired previously and decelerates the knowledge distillation. Therefore, the generator, apart from exploring the input space, seeks to compensate for the loss of knowledge in future iterations. Additionally, in a practical setting, during the distillation process, high variation in the student network's classification accuracy is undesirable, especially when the validation data is not available, since that prevents the user from tracking the student's accuracy over time, and selecting the distilled model parameters with the highest accuracy.
\begin{figure*}[t]
     \centering
     \begin{subfigure}[b]{0.33\textwidth}
         \centering
         \includegraphics[width=\textwidth]{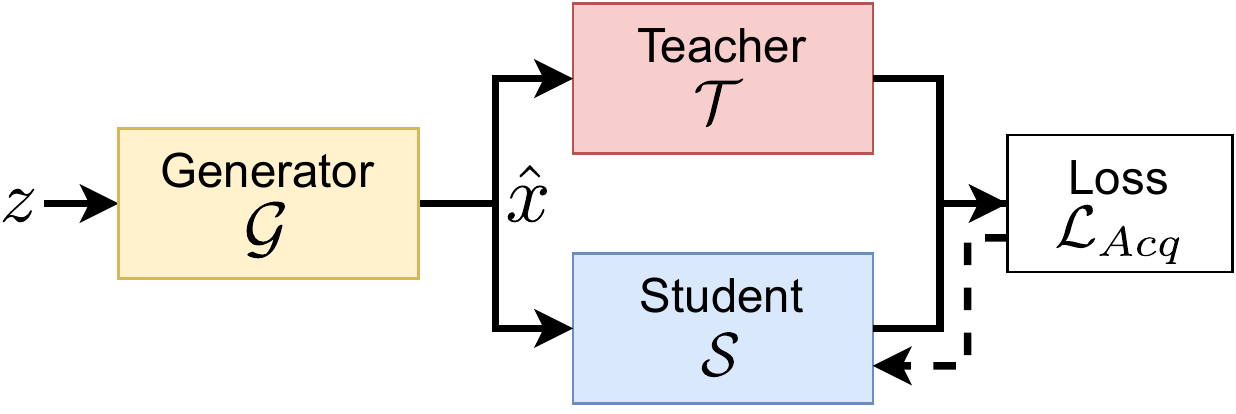}
         \caption{}
         \label{fig:overview_a}
     \end{subfigure}
     \hfill
     \begin{subfigure}[b]{0.33\textwidth}
         \centering
         \includegraphics[width=\textwidth]{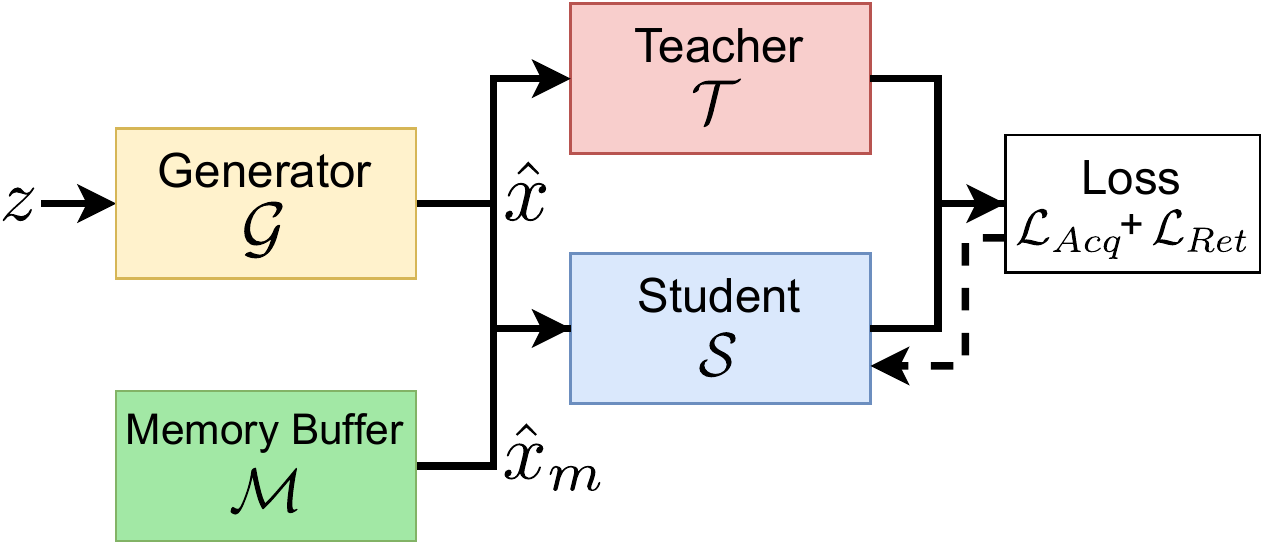}
         \caption{}
         \label{fig:overview_b}
     \end{subfigure}
     \hfill
     \begin{subfigure}[b]{0.33\textwidth}
         \centering
         \includegraphics[width=\textwidth]{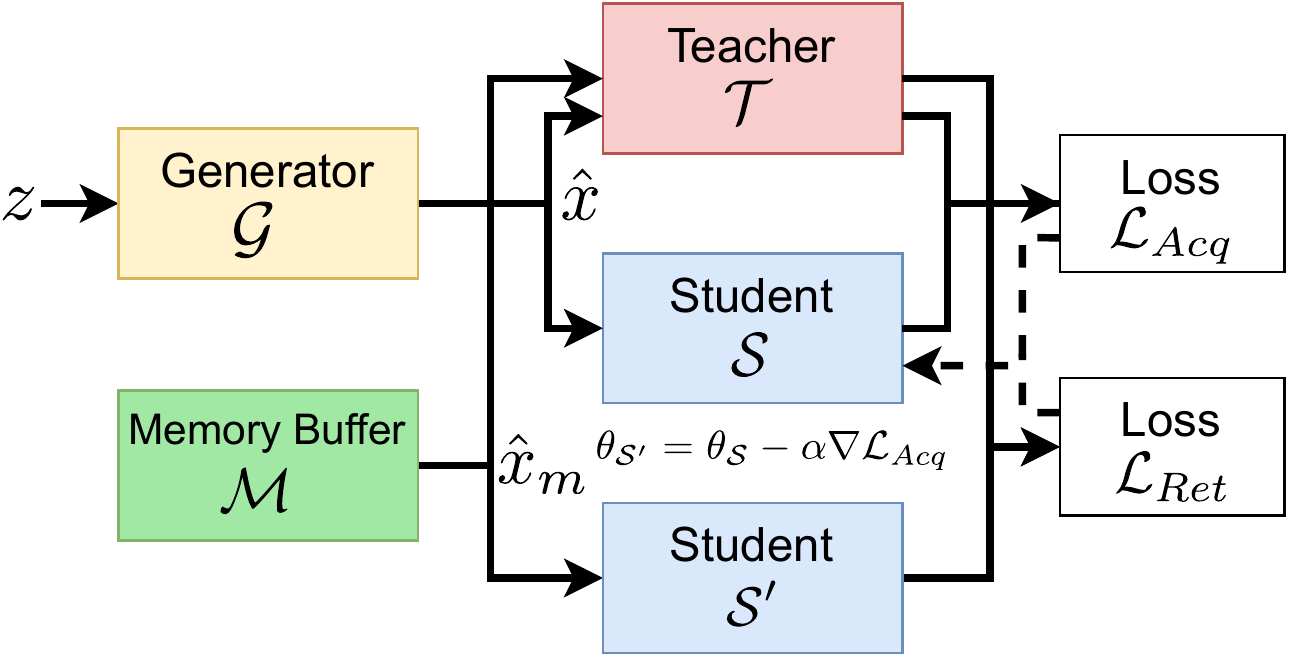}
         \caption{}
         \label{fig:overview_c}
     \end{subfigure}
        \caption{Student update strategies: (a) Typical student update by optimizing the \emph{Knowledge-Acquisition} loss ($\mathcal{L}_{Acq}$) with the batch pseudo samples ($\hat{x}$), produced by the generator ($\mathcal{G}$) \cite{fang2019data,micaelli2019zero,choi2020data}. (b) Student update with simultaneous optimization of the \emph{Knowledge-Acquisition}  loss ($\mathcal{L}_{Acq}$) and \emph{Knowledge-Retention} loss ($\mathcal{L}_{Ret}$) on the batch of pseduo samples ($\hat{x}$) and  memory samples ($\hat{x}_{m}$) obtained from the generator ($\mathcal{G}$) and the memory ($\mathcal{M}$), respectively \cite{binici2022preventing,binici2022robust}. (c) The proposed student update strategy, which treats $\mathcal{L}_{Acq}$ as meta-train and $\mathcal{L}_{Ret}$ as meta-test, and implicitly imposes the alignment between them.}
        \label{fig:overview}
\end{figure*}

To circumvent the above-discussed problem, existing methods maintain a memory buffer to rehearse the examples from previously encountered distributions while learning with current examples. Binci \textit{et al.} \cite{binici2022preventing} introduce \textit{Replay} based methods to explicitly retrain/replay on a  limited subset of previously encountered samples while training on the current examples. Then, carrying forward, they use Generative-Replay \cite{shin2017continual} to transfer the learned examples to an auxiliary generative model (VAE), and sample from the VAE's decoder in subsequent iterations \cite{binici2022robust}. Nonetheless, the performance of these methods is upper bounded by joint training on previous and current examples \cite{antiforgetting2}. Although, recent works have focused on modeling the memory, we seek to work towards effectively utilizing the samples from memory.

In this paper, we aim to update the student network parameters ($\theta_{\mathcal{S}}$) such that its performance does not degrade on the samples previously produced by the generator network ($\mathcal{G}$), aspiring towards \emph{Learning to Retain while Acquiring}. Thus, we propose a meta-learning inspired strategy to achieve this goal. We treat the task of \emph{Knowledge-Acquisition} (learning from newly generated samples) and \emph{Knowledge-Retention} (learning from previously encountered samples from memory) as meta-train and meta-test, respectively. Hence, in the proposed approach, the student network acquires new information while maintaining performance on previously encountered samples. By doing so, the proposed strategy (Figure \ref{fig:overview_c}) implicitly aligns \emph{Knowledge-Acquisition} and \emph{Knowledge-Retention}, as opposed to simply combining them \cite{binici2022preventing,binici2022robust} without any coordination or alignment (Figure \ref{fig:overview_b}), which leaves them to potentially interfere with one another. 

Additionally, analyzing the proposed meta-objective, we discover that (in Section \ref{sec:alignament_proof}) the latent alignment factor as the dot product between the gradients of the \emph{Knowledge-Acquisition} and \emph{Knowledge-Retention} objectives, suggesting that the meta-objective enforces a common gradient direction for both tasks, encouraging the alignment between the task-specific gradients. Thus, the proposed method simultaneously minimizes the loss and matches the gradients corresponding to the individual tasks (\emph{Knowledge-Acquisition} and \emph{Knowledge-Retention}), enforcing the optimization paths to be same for both tasks.

Moreover, the proposed student update strategy is scalable to different deep architectures as the gradient alignment is implicit, and memory-agnostic, making no assumptions about the replay scheme employed. Nonetheless, recent works on gradient alignment, have shown great empirical advantages in Zero-Shot Learning~\cite{ZeroGrad}, Distributed/Federated Learning~\cite{dandi2022implicit} and Domain Generalization~\cite{shi2022gradient}. Our method extends the advantages of gradient alignment to memory-based Adversarial DFKD, thus strengthening the empirical findings in these works. 

Finally, to demonstrate the advantages of the proposed student update strategy, we evaluate and compare against current non-memory~\cite{DAFL,fang2019data,choi2020data} and  memory-based~\cite{binici2022preventing,binici2022robust} Adversarial DFKD methods, and observe substantial improvement in the student learning evolution.

In summary, our contributions are as follows:
\begin{itemize}[noitemsep]
    \item We propose a novel meta-learning inspired student update strategy in the Adversarial DFKD setting, that aims to maintain the student’s performance on previously encountered examples (\emph{Knowledge-Retention}) while acquiring knowledge from samples of the current distribution (\emph{Knowledge-Acquisition}). 
    \item We theoretically identify (in Section \ref{sec:alignament_proof}) that the proposed student update strategy enforces an implicit gradient alignment between the \emph{Knowledge-Acquisition} and \emph{Knowledge-Retention} tasks.
    \item Finally, we evaluate our method and compare against various Adversarial DFKD methods, on multiple student architectures and replay schemes (Memory Buffer and Generative Replay).
\end{itemize}
\begin{figure*}[t]
    \centering
    \includegraphics[width=\textwidth]{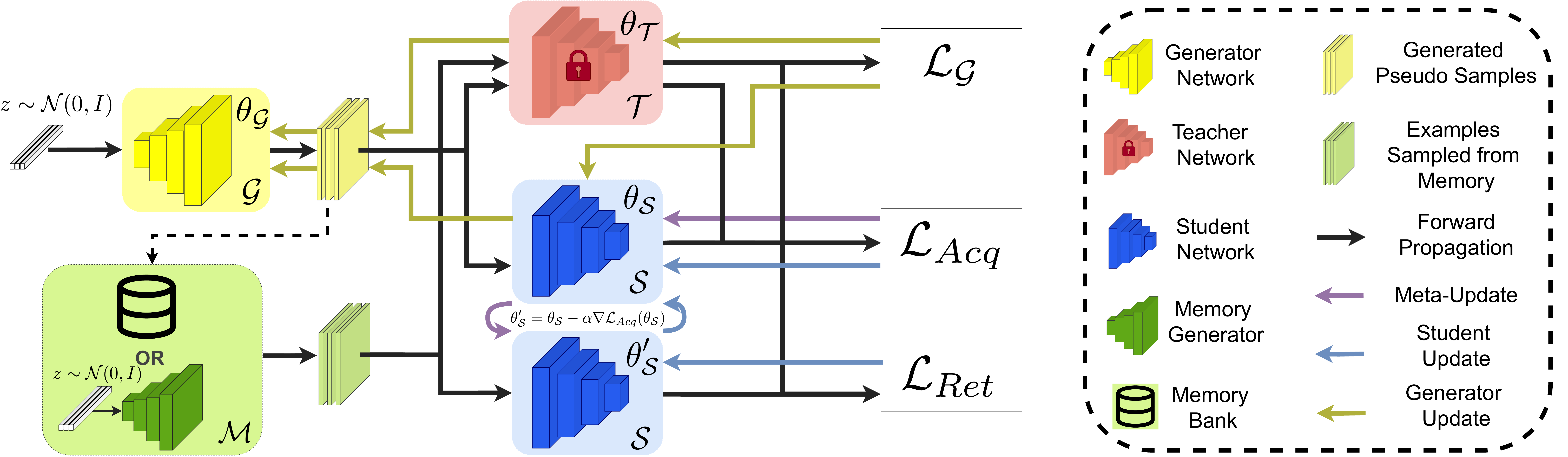}
    \caption{An illustration of the proposed DFKD framework. The framework consists of the Generator ($\mathcal{G}$), Teacher ($\mathcal{T}$), Student ($\mathcal{S}$) and the Memory ($\mathcal{M}$). $\mathcal{G}$ and $\mathcal{S}$ are updated alternatively, similar to the GAN \cite{GAN} framework with the generator loss ($\mathcal{L_{G}}$) optimizing $\mathcal{G}$, and the \emph{Knowledge-Acquisition} loss ($\mathcal{L}_{Acq}$) and the \emph{Knowledge-Retention} loss ($\mathcal{L}_{Ret}$) optimizing the student ($\mathcal{S}$). We use $\mathcal{M}$ in a generalized way to denote any type of replay schemes (Memory Buffer or Generative Replay in our case).}
    \label{fig:training_framework}
\end{figure*}
\section{Related Work}
\label{sec:related}
\noindent \textbf{Adversarial Data-Free Knowledge Distillation:} In the Adversarial Data-Free Knowledge Distillation paradigm, A generative model is trained to synthesize pseudo-samples that serve as queries for the Teacher ($\mathcal{T}$) and the Student $(\mathcal{S}$) \cite{choi2020data,micaelli2019zero,fang2019data}.  ZSKT \cite{micaelli2019zero} attempts data-free knowledge transfer by first training a generator in an adversarial fashion to look for samples on which the student and teacher do not match well. To improve the model discrepancy measure, it adopts the Kullback–Leibler (KL) divergence, and introduces attention transfer \cite{zagoruyko2017paying} to aid knowledge transfer. Moreover, DFAD \cite{fang2019data} recommends Mean Absolute Error (MAE) as a model discrepancy function to prevent decayed gradients on converged samples. Furthermore, the adversarial framework was extended by Choi \textit{et al.}~\cite{choi2020data} in the context of model quantization, by proposing adversarial data-free quantization (DFQ), and introducing additional regularization terms that match the mean and standard deviation of the generated pseudo-samples with the teacher model's batch-norm statistics, and imposes batch categorical entropy maximization, such that sample from each class appear equally in the generated batch. Fang \textit{et al.} recently introduced FastDFKD \cite{fang2022up}, an effective method with a meta generator to speed up the DFKD process, delivering a 100-fold increase in the knowledge transfer rate.\\\noindent \textbf{Handling Distribution Shift in Adversarial DFKD:} To counter the distribution mismatch and the catastrophic forgetting phenomenon in the adversarial framework \cite{seff2017continual}, Binici \textit{et al.} \cite{binici2022preventing} suggested maintaining a dynamic collection of synthetic samples throughout training iterations to prevent catastrophic forgetting in DFKD. Moreover, in their latest work \cite{binici2022robust}, they introduce generative pseudo-replay \cite{shin2017continual} in which an auxiliary generative model simultaneously learns the distribution of the samples produced by the generator ($\mathcal{G}$). Throughout the training process, examples are generated from the auxiliary generator to replay during training. Nonetheless, these works have focused on modeling the memory buffer. A related line of research maintains an exponentially moving average (EMA) of the generator model $\mathcal{G}$ \cite{do2022momentum} to replace the typical Memory-Buffer and Generative Replay. Nonetheless, our work focuses on the effective utilization of the samples obtained from the memory.

\section{Methodology}
\label{sec:method}
In Section \ref{sec:overview}, we first provide a brief overview of Adversarial DFKD. Then, in Section \ref{sec:goal}, we discuss the data-free knowledge distillation objective. In Sections \ref{sec:LRA} and \ref{sec:alignament_proof}, we elaborate on the proposed student update strategy. Lastly, in Section \ref{sec:generator}, we discuss the adopted generator update strategy used for the baselines and the proposed framework.

\subsection{Adversarial Data-Free Knowledge Distillation}
\label{sec:overview}
In the Adversarial DFKD framework, a generator ($\mathcal{G}$) is used to create pseudo-samples as surrogates to perform knowledge distillation/transfer, and the teacher-student ($\mathcal{T}$-$\mathcal{S}$) setup acts as a joint discriminator to penalize and update generator parameters ($\theta_{\mathcal{G}}$) in an adversarial manner. After updating $\theta_{\mathcal{G}}$, random samples are generated and used to minimize the $\mathcal{T}$-$\mathcal{S}$ discrepancy by updating the student parameters ($\theta_{\mathcal{S}}$). The generator and the student are optimized alternatively up until a fixed number of pre-defined iterations. In essence, the goal of DFKD is to craft a lightweight student model ($\mathcal{S}$) by harnessing valuable knowledge from the well-trained Teacher model ($\mathcal{T}$) in the absence of training data. A general overview of Adversarial DFKD framework is illustrated in Figure \ref{fig:intro}.
\begin{figure*}[t]
    \centering
    \includegraphics[width=\textwidth]{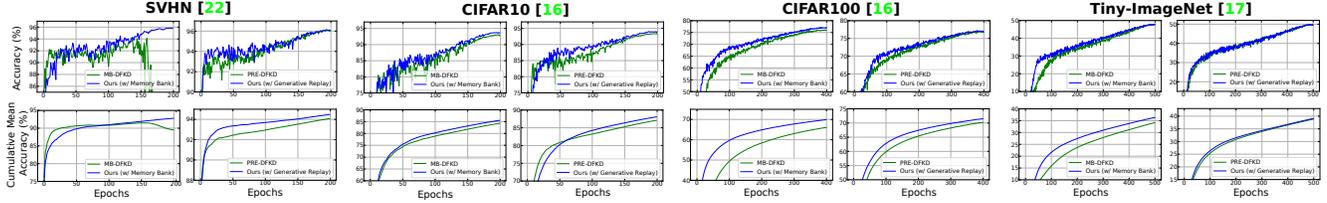}
    \caption{Learning evolution of the proposed method compared to the prior arts (MB-DFKD~\cite{binici2022preventing} and PRE-DFKD~\cite{binici2022robust}) that employ replay. The plots visualize the Accuracy (\%) evolution (\textit{top-row}) and the Cumulative Mean Accuracy (\%) evolution (\textit{bottom-row}) of the ResNet-18~\cite{he2016deep} student network on SVHN~\cite{SVHN}, CIFAR10~\cite{CIFAR}, CIFAR100~\cite{CIFAR}, and Tiny-ImageNet~\cite{tiny} datasets. The proposed method is in \textbf{\color[HTML]{3b80ee} Blue}.}
    \label{fig:learning_curves}
\end{figure*}
\subsection{Goal of Data Free Knowledge Distillation}
\label{sec:goal}
A student model ($\mathcal{S}$) is trained to match the teacher's ($\mathcal{T}$) predictions on its unavailable target domain ($\mathcal{D}_{\mathcal{T}}$) as part of the distillation process. Let, $p_{\mathcal{S}}(x) = \softmax(\mathcal{S}_{\theta_{\mathcal{S}}}(x))$ and $p_{\mathcal{T}}(x) = \softmax(\mathcal{T}_{\mathcal{\theta_{T}}}(x))$, $\forall x  \in \mathcal{D}_{\mathcal{T}}$, denote the predicted Student and Teacher probability mass across the classes, respectively. We seek to find the parameters $\theta_{\mathcal{S}}$ of the student model that will reduce the probability of errors ($P$) between the predictions $p_{\mathcal{S}}(x)$ and $p_{\mathcal{T}}(x)$:
\begin{equation}
    \min_{\theta_{\mathcal{S}}}P_{x \sim \mathcal{D}_{\mathcal{T}}}\left(  \argmax_{i}{p_{\mathcal{S}}^{i}(x)} \neq \argmax_{i}{p_{\mathcal{T}}^{i}(x)}  \right), \nonumber
\end{equation}
where the superscript $i$ denotes the $i^{th}$ probability score of the predicted masses, $p_{\mathcal{S}}(x)$ and $p_{\mathcal{T}}(x)$. 

However, DFKD suggests minimizing the student's error on a pseudo dataset $\mathcal{D}_{\mathcal{P}}$, since the teacher's original training data distribution $\mathcal{D}_{\mathcal{T}}$ is not available. Typically, a loss function, say $\mathcal{L}_{KD}$, that gauges disagreement between the teacher and the student is minimized  $\forall \hat{x} \in \mathcal{D}_{\mathcal{P}}$ by optimizing the student parameters ($\theta_{\mathcal{S}}$):
\begin{equation}
    \min_{\theta_{S}}{\mathbb{E}_{\hat{x}\sim\mathcal{D}_{\mathcal{P}}}[\mathcal{L}_{KD}(\mathcal{T}_{\theta_{\mathcal{T}}}(\hat{x}), \mathcal{S}_{\theta_{\mathcal{S}}}(\hat{x}))]}.
\end{equation}
We use the Mean Absolute Error (MAE) to define loss measure ($\mathcal{L}_{KD}$) as suggested in \cite{fang2019data}. Suppose, given the predicted logits (pre-softmax predictions) $t(\hat{x}) = \mathcal{T}_{\theta_{\mathcal{T}}}(\hat{x})$ from the teacher model and the predicted logits $s(\hat{x})=\mathcal{S}_{\theta_{\mathcal{S}}}(\hat{x})$ from the student model, we define $\mathcal{L}_{KD}$  as:
\begin{equation}
    \mathcal{L}_{KD}(t(\hat{x}), s(\hat{x})) = \lVert t(\hat{x}) - s(\hat{x}) \rVert_{1}.
    \label{eq:mae}
\end{equation}
\subsection{Learning to Retain while Acquiring}
\label{sec:LRA}
Our novelty resides in the student update strategy, as described earlier. \emph{Knowledge-Retention} and \emph{Knowledge-Acquisition} relate to two independent goals and can therefore be considered as two discrete tasks. In fact, by aligning these two goals, we empirically observe that, they can cooperate to retain the knowledge on previously encountered examples while acquiring knowledge from newly generated samples. The proposed method utilizes a meta-learning-inspired optimization strategy to effectively replay the samples from memory, say $\mathcal{M}$.

In the typical Adversarial DFKD setup, the student update objective with the generated pseudo samples ($\hat{x}$), \textit{i.e.}, the \emph{Knowledge-Acquisition} task ($\mathcal{L}_{Acq}$) (Figure \ref{fig:overview_a}), is formulated as:
\begin{equation}
    \min_{\theta_{\mathcal{S}}}\mathcal{L}_{Acq}(\theta_{\mathcal{S}}) = \min_{\theta_{\mathcal{S}}}
    \mathbb{E}_{\hat{x}}[\mathcal{L}(\mathcal{T}_{\theta_{\mathcal{T}}}(\hat{x}), \mathcal{S}_{\theta_{\mathcal{S}}}(\hat{x}))],
\end{equation}
where $\mathcal{L}$ is the MAE (\ref{eq:mae}) between the teacher and the student logits, and $\hat{x} = \mathcal{G}(z), z\sim \mathcal{N}(0,I)$, denotes the batch of randomly generated samples.

Moreover, to alleviate the distribution drift during knowledge distillation in the adversarial setting, previous works have maintained a memory buffer, to store $\hat{x}$ and replay at later iterations to help the student recall the knowledge \cite{binici2022preventing}. At a fixed frequency, batches of samples from current iterations are updated in the memory \cite{binici2022preventing}. Also, recent works~\cite{binici2022robust} have explored using the Generative Replay~\cite{shin2017continual} strategy to simultaneously store the samples, as they are encountered, in a generative model (VAE~\cite{VAE}), and later replay by generating samples from the VAE decoder. Hence, the optimization objective for the \emph{Knowledge-Retention} ($\mathcal{L}_{Ret}$) can be defined as follows:
 \begin{align}
     \min_{\theta_{\mathcal{S}}} \mathcal{L}_{Ret}(\theta_{\mathcal{S}}) = \min_{\theta_{\mathcal{S}}}\mathbb{E}_{\hat{x}_{m}}[\mathcal{L}(\mathcal{T}_{\theta_{\mathcal{T}}}(\hat{x}_{m}), \mathcal{S}_{\theta_{\mathcal{S}}}(\hat{x}_{m}))], \nonumber \\
     \hat{x}_{m} \sim \mathcal{M},
 \end{align}
 where, with a slight abuse of notation, $\mathcal{M}$ denotes a Memory Buffer or Generative Replay or any other replay scheme. Thus, the overall optimization objective to update $\mathcal{\theta_{\mathcal{S}}}$, while considering both the new samples (generated from the latest $\theta_\mathcal{G}$) and the old samples (sampled from $\mathcal{M}$) (Figure \ref{fig:overview_b}) is defined as:
 \begin{equation}
 \label{eq:naive_replay}
     \min_{\theta_{\mathcal{S}}}\mathcal{L}_{Acq}(\theta_{\mathcal{S}}) + \mathcal{L}_{Ret}(\theta_{\mathcal{S}}).
 \end{equation}
 
However, the objective in (\ref{eq:naive_replay}) attempts to simultaneously optimizes $\mathcal{L}_{Ret}$ and $\mathcal{L}_{Acq}$ but does not seek to align the objectives, which leaves them to potentially interfere with one another. Say, we denote the gradients of \emph{Knowledge Acquisition} and \emph{Knowledge Retention} tasks with $\nabla \mathcal{L}_{Acq}(\theta)$ and $\nabla \mathcal{L}_{Ret}(\theta)$, respectively. If $\nabla \mathcal{L}_{Acq}(\theta)$ and $\nabla \mathcal{L}_{Ret}(\theta)$ point in a similar direction or are said to be aligned, \textit{i.e.}, $\nabla \mathcal{L}_{Acq}(\theta).\nabla \mathcal{L}_{Ret}(\theta) > 0$, then taking a gradient step along $\nabla \mathcal{L}_{Acq}(\theta)$ or $\nabla \mathcal{L}_{Ret}(\theta)$ improves the model’s performance on both tasks. This is however not the case when $\nabla \mathcal{L}_{Acq}(\theta)$ and  $\nabla \mathcal{L}_{Ret}(\theta)$ point in different directions, \textit{i.e.}, $\nabla \mathcal{L}_{Acq}(\theta).\nabla \mathcal{L}_{Ret}(\theta)\leq 0$, and the weight parameters obtained, may not be optimal for both the tasks simultaneously. Hence, the intended effect of having the gradient directions aligned is to obtain student parameters ($\theta_{\mathcal{S}}$) that have optimal performance on both $\mathcal{L}_{Acq}$ and $\mathcal{L}_{Ret}$.

The proposed meta-learning inspired approach, seeks to align the two tasks, $\mathcal{L}_{Acq}$ and $\mathcal{L}_{Ret}$. We take cues from Model-Agnostic Meta-learning (MAML)\cite{finn2017model}, and adapt it to Adversarial DFKD. At each iteration of the parameter update, we pose \emph{Knowledge-Acquisition} and \emph{Knowledge-Acquisition} as meta-train and meta-test, respectively. Which means, we perform a single gradient descent step using the current samples ($\hat{x}$) produced from the generator network ($\mathcal{G}$), on the parameters $\mathcal{\theta_{S}}$  and obtain $\mathcal{\theta_{S}}^{\prime}$ as $\theta_{\mathcal{S}}^\prime = \theta_{\mathcal{S}} - \alpha \nabla \mathcal{L}_{Acq}(\theta_{\mathcal{S}})$, where $\nabla \mathcal{L}_{Acq}(\theta_{\mathcal{S}})$ denotes gradient of $\mathcal{L}_{Acq}$ at $\theta_{\mathcal{S}}$. Then we optimize on the batch of samples ($\hat{x}_{m}$) obtained from the memory ($\mathcal{M}$),  with the parameters $\theta_{\mathcal{S}}^\prime$, and then make a final update on $\theta_{\mathcal{S}}$. Thus, formally, the overall meta-optimization objective, with the task of \emph{Knowledge Acquisition} serving as meta-train and the task of \emph{Knowledge Retention} as meta-test (Figure \ref{fig:overview_c}), can be defined as follows:
\begin{align}
\label{eq:meta_objective}
    \min_{\theta_{\mathcal{S}}} \mathcal{L}_{Acq}(\theta_{\mathcal{S}}) + \mathcal{L}_{Ret}(\theta_{\mathcal{S}}^{\prime}) &= \min_{\theta_{\mathcal{S}}} \mathcal{L}_{Acq}(\theta_{\mathcal{S}}) + \nonumber \\ & \mathcal{L}_{Ret}(\theta_{\mathcal{S}} - \alpha \nabla \mathcal{L}_{Acq}(\theta_{\mathcal{S}})). 
\end{align}
\subsection{How does the Proposed Student Update Strategy Promote Alignment?}\label{sec:alignament_proof}
In this subsection, we analyze the proposed objective (\ref{eq:meta_objective}) to understand how it results in the desired alignment between the $\emph{Knowledge-Acquisiton}$ and $\emph{Knowledge-Retention}$ tasks. We assume that meta-train and meta-test tasks provide us with losses $\mathcal{L}_{Acq}$ and $\mathcal{L}_{Ret}$; in our case, $\mathcal{L}_{Acq}$ and $\mathcal{L}_{Ret}$ are the same function computed on the batches $\hat{x}$ and $\hat{x}_{m}$, respectively. We utilize Taylor's expansion to elaborate the gradient of $\mathcal{L}_{Ret}$ at a point $\theta$ displaced by $\phi_{\theta}$, as described in Lemma \ref{lemma:taylor_series}, 
\begin{customlemma}{1}\label{lemma:taylor_series}
If $\mathcal{L}_{Ret}$ has Lipschitz Hessian, \textit{i.e.}, $\lVert \nabla^{2}\mathcal{L}_{Ret}(\theta_{1}) - \nabla^{2}\mathcal{L}_{Ret}(\theta_{2}) \rVert \leq \rho \lVert \theta_{1} - \theta_{2} \rVert$ for some $\rho > 0$, then:
\begin{align*}
    \nabla\mathcal{L}_{Ret}(\theta + \mathbf{\phi}_{\theta}) = \nabla\mathcal{L}_{Ret}(\theta) +  \nabla^{2}\mathcal{L}_{Ret}(\theta)\mathbf{\phi}_{\theta}\\ + \mathcal{O}(\lVert \mathbf{\phi}_{\theta} \rVert^{2}).
\end{align*}
For instance, when $\phi_{\theta} = -\alpha \nabla \mathcal{L}_{Acq}(\theta)$, we have,
\begin{align*}
    \nabla\mathcal{L}_{Ret}(\theta -\alpha \nabla \mathcal{L}_{Acq}(\theta)) = & \nabla\mathcal{L}_{Ret}(\theta)\\ &-  \alpha \nabla^{2}\mathcal{L}_{Ret}(\theta) \nabla \mathcal{L}_{Acq}(\theta)\\ &+ \mathcal{O}(\alpha^{2}). \end{align*}
\end{customlemma}
\begin{customthm}{1}\label{theorem:1}
If $\theta^{\prime} = \theta - \alpha\nabla \mathcal{L}_{Acq}(\theta)$, denotes a single gradient descent step on $\theta$ with the objective $\mathcal{L}_{Acq}(\theta)$, where $\alpha$ is a scalar, and $\nabla \mathcal{L}_{Acq}(\theta)$ denotes the gradient of the objective at $\theta$, then:
\begin{align*}
    \frac{\partial \mathcal{L}_{Ret}(\theta^{\prime})}{\partial \theta} = \nabla \mathcal{L}_{Ret}(\theta) - \alpha \nabla^{2} \mathcal{L}_{Ret}(\theta).\nabla \mathcal{L}_{Acq}(\theta)\\ - \alpha \nabla^{2} \mathcal{L}_{Acq}(\theta).\nabla \mathcal{L}_{Ret}(\theta) + \mathcal{O}(\alpha^{2}).
\end{align*}
\end{customthm}
\begin{proof}\let\qed\relax
Please refer to the Supplemental Material (Theorem \ref{lemma:theorem1_proof}. \nonumber
\end{proof}
\noindent While optimizing the objective defined in (\ref{eq:meta_objective}) using stochastic gradient descent, we would need to compute the gradient of the $\mathcal{L}_{Ret}(\theta_{\mathcal{S}}^{\prime})$ w.r.t $\theta_{\mathcal{S}}$. Therefore, utilizing Theorem \ref{theorem:1} we express $\frac{\partial \mathcal{L}_{Ret}(\theta^{\prime}_{\mathcal{S}})}{\partial \theta_{\mathcal{S}}}$ as:
\begin{align}
\frac{\partial \mathcal{L}_{Ret}(\theta^{\prime}_{\mathcal{S}})}{\partial \theta_{\mathcal{S}}} &= \nabla \mathcal{L}_{Ret}(\theta_{\mathcal{S}}) \nonumber \\ &- \alpha \nabla^{2}\mathcal{L}_{Ret}(\theta_{\mathcal{S}}).\nabla \mathcal{L}_{Acq}(\theta_{\mathcal{S}}) \nonumber \\ &- \alpha \nabla^{2} \mathcal{L}_{Acq}(\theta_{\mathcal{S}}).\nabla \mathcal{L}_{Ret}(\theta_{\mathcal{S}}) + \mathcal{O}(\alpha^{2}),
\end{align}
\noindent using the product rule $\nabla a .b + \nabla b.a = \nabla(a.b)$, we get:
\begin{align}
\label{eq:grad_align}
\frac{\partial \mathcal{L}_{Ret}(\theta^{\prime}_{\mathcal{S}})}{\partial \theta_{\mathcal{S}}} & = \nabla \mathcal{L}_{Ret}(\theta_{\mathcal{S}}) \nonumber \\
&- \alpha \nabla \underbrace{(\nabla \mathcal{L}_{Ret}(\theta_{\mathcal{S}}).\nabla \mathcal{L}_{Acq}(\theta_{\mathcal{S}}))}_{Gradient \ Alignment} + \mathcal{O}(\alpha^{2}). 
\end{align}
From the analysis above, we observe that the gradient of $\mathcal{L}_{Ret}(\theta_{\mathcal{S}}^{\prime})$ at $\theta_{\mathcal{S}}$ (in (\ref{eq:grad_align})) produces the gradient of the gradient-product. This indicates, when optimizing $\mathcal{L}_{Ret}(\theta^{\prime}_{\mathcal{S}})$ (in (\ref{eq:meta_objective})), the gradient of $\mathcal{L}_{Ret}(\theta_{\mathcal{S}}^{\prime})$ at $\theta_{\mathcal{S}}$, has a negatively scaled gradient of the gradient-product term $ \nabla (\nabla \mathcal{L}_{Ret}(\theta_{\mathcal{S}}).\nabla \mathcal{L}_{Acq}(\theta_{\mathcal{S}}))$ (derived in (\ref{eq:grad_align})), suggesting that the overall-gradients minimize $\mathcal{L}_{Ret}(\theta_{\mathcal{S}})$ and maximize $\nabla \mathcal{L}_{Ret}(\theta_{\mathcal{S}}).\nabla \mathcal{L}_{Acq}(\theta_{\mathcal{S}})$. Hence, optimizing (\ref{eq:meta_objective}) enforces the updates on $\mathcal{L}_{Ret}(\theta_{\mathcal{S}})$ and $\mathcal{L}_{Acq}(\theta_{\mathcal{S}})$ to seek a common direction, by maximizing the gradient-product.
\begin{table*}[ht]
\centering
\caption{Distillation results of the Adversarial DFKD methods on four image classification benchmark datasets, SVHN \cite{SVHN}, CIFAR10 \cite{CIFAR}, CIFAR100 \cite{CIFAR}, Tiny-ImageNet \cite{tiny}. Primarily we compare against the replay-based methods, present at the bottom panel of each dataset. The best numbers from our method on the Bank-based replay are highlighted in \textbf{{\color[HTML]{00009B}Navy}}, and for the one with Generative replay are highlighted in \textbf{{\color[HTML]{9A0000}Maroon}}. The first five columns represent the $\mu[\mathcal{S}_{Acc}]$ and the $\sigma^{2}[\mathcal{S}_{Acc}]$ at different epoch percentiles (described in Section \ref{sec:experiments}). The last column represents the maximum test accuracy (Acc\textsubscript{max} (\%)) attained by the student network in the entire training period. Note: For SVHN and Tiny-ImageNet we were unable to reproduce the results of DFQ \cite{choi2020data} in our setting. DFKD\textsuperscript{*} denotes the baseline method without any replay. }
\resizebox{0.9\textwidth}{!}{%
\begin{tabular}{@{}lccccccccccc@{}}
\toprule
\multicolumn{1}{c}{}                                                                                             & \multicolumn{2}{c}{$\mathbf{>0^{th} \: \text{\textbf{Percentile}}} $}                                                            & \multicolumn{2}{c}{$\mathbf{>20^{th}\: \text{\textbf{Percentile}}}$}                                                             & \multicolumn{2}{c}{$\mathbf{>40^{th}\: \text{\textbf{Percentile}}}$}                                                             & \multicolumn{2}{c}{$\mathbf{>60^{th}\: \text{\textbf{Percentile}}}$}                                                            & \multicolumn{2}{c}{$\mathbf{>80^{th}\: \text{\textbf{Percentile}}}$}                                                            &                                                        \\ \cmidrule(lr){2-11}
\multicolumn{1}{c}{\multirow{-2}{*}{\textbf{Method}}}                                                            & $\mu \uparrow$                              & \multicolumn{1}{c|}{$\sigma^{2} \downarrow$}                                       & $\mu \uparrow$                              & \multicolumn{1}{c|}{$\sigma^{2} \downarrow$}                                       & $\mu \uparrow$                              & \multicolumn{1}{c|}{$\sigma^{2} \downarrow$}                                       & $\mu \uparrow$                              & \multicolumn{1}{c|}{$\sigma^{2} \downarrow$}                                      & $\mu \uparrow$                              & $\sigma^{2} \downarrow$                                                           & \multirow{-2}{*}{\textbf{Acc\textsubscript{max} (\%)}} \\ \midrule
\multicolumn{12}{c}{\textbf{SVHN \cite{SVHN}}}                                                                                                                                                                                                                                                                                                                                                                                                                                                                                                                                                                                                                                                                                                                                                                                                                     \\
\multicolumn{12}{c}{$\boldsymbol{\mathcal{T}_{Acc}  = 97.45\%}$, $\boldsymbol{\mathcal{S}_{Acc} = 97.26\%}$, \textbf{ ($\mathcal{E}_{max}$)}  = \textbf{200}}                                                                                                                                                                                                                                                                                                                                                                                                                                                                                                                                                                                                                                                                                                                                      \\ \midrule
\multicolumn{1}{l|}{{\color[HTML]{343434}DAFL \cite{DAFL}}}                                                                 & {\color[HTML]{343434} 85.15}                & \multicolumn{1}{c|}{{\color[HTML]{343434} 241.70}}                                 & {\color[HTML]{343434} 90.49}                & \multicolumn{1}{c|}{{\color[HTML]{343434} 8.71}}                                   & {\color[HTML]{343434} 91.75}                & \multicolumn{1}{c|}{{\color[HTML]{343434} 3.73}}                                   & {\color[HTML]{343434} 92.88}                & \multicolumn{1}{c|}{{\color[HTML]{343434} 0.84}}                                  & {\color[HTML]{343434} 93.59}                & \multicolumn{1}{c|}{{\color[HTML]{343434} 0.20}}                                  & {\color[HTML]{343434} 94.23}                           \\
\multicolumn{1}{l|}{{\color[HTML]{343434} DFAD \cite{fang2019data}}}                                                                 & {\color[HTML]{343434} 92.52}                & \multicolumn{1}{c|}{{\color[HTML]{343434} 93.42}}                                  & {\color[HTML]{343434} 94.10}                & \multicolumn{1}{c|}{{\color[HTML]{343434} 0.07}}                                   & {\color[HTML]{343434} 94.13}                & \multicolumn{1}{c|}{{\color[HTML]{343434} 0.07}}                                   & {\color[HTML]{343434} 94.14}                & \multicolumn{1}{c|}{{\color[HTML]{343434} 0.08}}                                  & {\color[HTML]{343434} 94.11}                & \multicolumn{1}{c|}{{\color[HTML]{343434} 0.07}}                                  & {\color[HTML]{343434} 94.68}                           \\
\multicolumn{1}{l|}{{\color[HTML]{343434} DFKD\textsuperscript{*} \cite{binici2022preventing}}}                                                                 & {\color[HTML]{343434} 91.84}                & \multicolumn{1}{c|}{{\color[HTML]{343434} 8.59}}                                   & {\color[HTML]{343434} 92.53}                & \multicolumn{1}{c|}{{\color[HTML]{343434} 6.76}}                                   & {\color[HTML]{343434} 93.43}                & \multicolumn{1}{c|}{{\color[HTML]{343434} 4.39}}                                   & {\color[HTML]{343434} 94.61}                & \multicolumn{1}{c|}{{\color[HTML]{343434} 1.72}}                                  & {\color[HTML]{343434} 95.72}                & \multicolumn{1}{c|}{{\color[HTML]{343434} 0.04}}                                  & {\color[HTML]{343434} 95.97}                           \\ \midrule
\multicolumn{1}{l|}{{\color[HTML]{000000} MB-DFKD \cite{binici2022preventing}}}                                                              & {\color[HTML]{000000} 89.51}                & \multicolumn{1}{c|}{{\color[HTML]{000000} 21.13}}                                  & {\color[HTML]{000000} 89.28}                & \multicolumn{1}{c|}{{\color[HTML]{000000} 24.31}}                                  & {\color[HTML]{000000} 88.65}                & \multicolumn{1}{c|}{{\color[HTML]{000000} 30.63}}                                  & {\color[HTML]{000000} 87.14}                & \multicolumn{1}{c|}{{\color[HTML]{000000} 38.67}}                                 & {\color[HTML]{000000} 81.67}                & \multicolumn{1}{c|}{{\color[HTML]{000000} 15.24}}                                 & {\color[HTML]{000000} 94.11}                           \\
\multicolumn{1}{l|}{{\color[HTML]{000000} PRE-DFKD \cite{binici2022robust}}}                                                             & {\color[HTML]{000000} 94.08}                & \multicolumn{1}{c|}{{\color[HTML]{000000} 2.11}}                                   & {\color[HTML]{000000} 94.53}                & \multicolumn{1}{c|}{{\color[HTML]{000000} 1.11}}                                   & {\color[HTML]{000000} 94.97}                & \multicolumn{1}{c|}{{\color[HTML]{000000} 0.66}}                                   & {\color[HTML]{000000} 95.42}                & \multicolumn{1}{c|}{{\color[HTML]{9A0000} \textbf{0.26}}}                                  & {\color[HTML]{000000} 95.85}                & \multicolumn{1}{c|}{{\color[HTML]{000000} 0.04}}                                  & {\color[HTML]{000000} 96.10}                           \\
\rowcolor[HTML]{EFEFEF} 
\multicolumn{1}{l|}{\cellcolor[HTML]{EFEFEF}{\color[HTML]{000000} \textit{\textbf{Ours-1 (Memory Bank)}}}}       & {\color[HTML]{00009B} {\textbf{92.78}}} & \multicolumn{1}{c|}{\cellcolor[HTML]{EFEFEF}{\color[HTML]{00009B} \textbf{7.44}}}  & {\color[HTML]{00009B} { \textbf{93.69}}} & \multicolumn{1}{c|}{\cellcolor[HTML]{EFEFEF}{\color[HTML]{00009B} \textbf{2.21}}}  & {\color[HTML]{00009B} {\textbf{94.25}}} & \multicolumn{1}{c|}{\cellcolor[HTML]{EFEFEF}{\color[HTML]{00009B} \textbf{1.54}}}  & {\color[HTML]{00009B} { \textbf{94.94}}} & \multicolumn{1}{c|}{\cellcolor[HTML]{EFEFEF}{\color[HTML]{00009B} \textbf{0.62}}} & {\color[HTML]{00009B} { \textbf{95.59}}} & \multicolumn{1}{c|}{\cellcolor[HTML]{EFEFEF}{\color[HTML]{00009B} \textbf{0.05}}} & {\color[HTML]{00009B} \textbf{95.88}}                  \\
\rowcolor[HTML]{EFEFEF} 
\multicolumn{1}{l|}{\cellcolor[HTML]{EFEFEF}{\color[HTML]{000000} \textit{\textbf{Ours-2 (Generative Replay)}}}} & {\color[HTML]{9A0000} {\textbf{94.48}}} & \multicolumn{1}{c|}{\cellcolor[HTML]{EFEFEF}{\color[HTML]{9A0000} \textbf{1.64}}}  & {\color[HTML]{9A0000} { \textbf{94.79}}} & \multicolumn{1}{c|}{\cellcolor[HTML]{EFEFEF}{\color[HTML]{9A0000} \textbf{0.71}}}  & {\color[HTML]{9A0000} { \textbf{95.09}}} & \multicolumn{1}{c|}{\cellcolor[HTML]{EFEFEF}{\color[HTML]{9A0000} \textbf{0.51}}}  & {\color[HTML]{9A0000} { \textbf{95.47}}} & \multicolumn{1}{c|}{\cellcolor[HTML]{EFEFEF}{\color[HTML]{000000} 0.27}}          & {\color[HTML]{9A0000} { \textbf{95.91}}} & \multicolumn{1}{c|}{\cellcolor[HTML]{EFEFEF}{\color[HTML]{9A0000} \textbf{0.03}}} & {\color[HTML]{9A0000} \textbf{96.15}}                  \\ \midrule
\multicolumn{12}{c}{\textbf{CIFAR10 \cite{CIFAR}}}                                                                                                                                                                                                                                                                                                                                                                                                                                                                                                                                                                                                                                                                                                                                                                                                                  \\
\multicolumn{12}{c}{$\boldsymbol{\mathcal{T}_{Acc} = 95.72\%}$, $\boldsymbol{\mathcal{S}_{Acc} = 95.23\%}$, \textbf{ ($\mathcal{E}_{max}$)}  = \textbf{200}}                                                                                                                                                                                                                                                                                                                                                                                                                                                                                                                                                                                                                                                                                                                                      \\ \midrule
\multicolumn{1}{l|}{{\color[HTML]{343434} DAFL \cite{DAFL}}}                                                                 & {\color[HTML]{343434} 62.78}                & \multicolumn{1}{c|}{{\color[HTML]{343434} 629.13}}                                 & {\color[HTML]{343434} 72.51}                & \multicolumn{1}{c|}{{\color[HTML]{343434} 296.44}}                                 & {\color[HTML]{343434} 80.57}                & \multicolumn{1}{c|}{{\color[HTML]{343434} 110.18}}                                 & {\color[HTML]{343434} 87.05}                & \multicolumn{1}{c|}{{\color[HTML]{343434} 22.40}}                                 & {\color[HTML]{343434} 90.76}                & \multicolumn{1}{c|}{{\color[HTML]{343434} 2.04}}                                  & {\color[HTML]{343434} 92.23}                           \\
\multicolumn{1}{l|}{{\color[HTML]{343434} DFAD \cite{fang2019data}}}                                                                 & {\color[HTML]{343434} 85.88}                & \multicolumn{1}{c|}{{\color[HTML]{343434} 152.39}}                                 & {\color[HTML]{343434} 89.91}                & \multicolumn{1}{c|}{{\color[HTML]{343434} 11.95}}                                  & {\color[HTML]{343434} 91.46}                & \multicolumn{1}{c|}{{\color[HTML]{343434} 5.66}}                                   & {\color[HTML]{343434} 92.66}                & \multicolumn{1}{c|}{{\color[HTML]{343434} 0.07}}                                  & {\color[HTML]{343434} 92.85}                & \multicolumn{1}{c|}{{\color[HTML]{343434} 0.04}}                                  & {\color[HTML]{343434} 93.21}                           \\
\multicolumn{1}{l|}{{\color[HTML]{343434} DFQ \cite{choi2020data}}}                                                                  & {\color[HTML]{343434} 81.25}                & \multicolumn{1}{c|}{{\color[HTML]{343434} 98.86}}                                  & {\color[HTML]{343434} 84.41}                & \multicolumn{1}{c|}{{\color[HTML]{343434} 37.40}}                                  & {\color[HTML]{343434} 87.08}                & \multicolumn{1}{c|}{{\color[HTML]{343434} 19.59}}                                  & {\color[HTML]{343434} 89.72}                & \multicolumn{1}{c|}{{\color[HTML]{343434} 6.15}}                                  & {\color[HTML]{343434} 91.87}                & \multicolumn{1}{c|}{{\color[HTML]{343434} 0.67}}                                  & {\color[HTML]{343434} 92.90}                           \\
\multicolumn{1}{l|}{{\color[HTML]{343434} DFKD\textsuperscript{*} \cite{binici2022preventing}}}                                                                 & {\color[HTML]{343434} 83.57}                & \multicolumn{1}{c|}{{\color[HTML]{343434} 107.77}}                                 & {\color[HTML]{343434} 86.94}                & \multicolumn{1}{c|}{{\color[HTML]{343434} 18.42}}                                  & {\color[HTML]{343434} 88.67}                & \multicolumn{1}{c|}{{\color[HTML]{343434} 11.83}}                                  & {\color[HTML]{343434} 90.67}                & \multicolumn{1}{c|}{{\color[HTML]{343434} 4.09}}                                  & {\color[HTML]{343434} 92.32}                & \multicolumn{1}{c|}{{\color[HTML]{343434} 0.38}}                                  & {\color[HTML]{343434} 93.02}                           \\ \midrule
\multicolumn{1}{l|}{{\color[HTML]{000000} MB-DFKD \cite{binici2022preventing}}}                                                              & {\color[HTML]{000000} 84.29}                & \multicolumn{1}{c|}{{\color[HTML]{000000} 95.74}}                                  & {\color[HTML]{000000} 87.25}                & \multicolumn{1}{c|}{{\color[HTML]{000000} 16.47}}                                  & {\color[HTML]{000000} 88.81}                & \multicolumn{1}{c|}{{\color[HTML]{000000} 11.00}}                                  & {\color[HTML]{000000} 90.71}                & \multicolumn{1}{c|}{{\color[HTML]{000000} 3.79}}                                  & {\color[HTML]{000000} 92.31}                & \multicolumn{1}{c|}{{\color[HTML]{000000} 0.35}}                                  & {\color[HTML]{000000} 93.03}                           \\
\multicolumn{1}{l|}{{\color[HTML]{000000} PRE-DFKD \cite{binici2022robust}}}                                                             & {\color[HTML]{000000} 87.10}                & \multicolumn{1}{c|}{\textbf{\color[HTML]{9A0000} 45.69}}                                  & {\color[HTML]{000000} 88.88}                & \multicolumn{1}{c|}{{\color[HTML]{000000} 10.86}}                                  & {\color[HTML]{000000} 90.24}                & \multicolumn{1}{c|}{{\color[HTML]{000000} 6.26}}                                   & {\color[HTML]{000000} 91.75}                & \multicolumn{1}{c|}{{\color[HTML]{000000} 1.84}}                                  & {\color[HTML]{000000} 92.92}                & \multicolumn{1}{c|}{{\color[HTML]{9A0000}\textbf{0.14}}}                                  & {\color[HTML]{000000} 93.41}                           \\
\rowcolor[HTML]{EFEFEF} 
\multicolumn{1}{l|}{\cellcolor[HTML]{EFEFEF}{\color[HTML]{000000} \textit{\textbf{Ours-1 (Memory Bank)}}}}       & {\color[HTML]{00009B} {\ \textbf{85.53}}} & \multicolumn{1}{c|}{\cellcolor[HTML]{EFEFEF}{\color[HTML]{00009B} \textbf{91.33}}} & {\color[HTML]{00009B} {\textbf{88.58}}} & \multicolumn{1}{c|}{\cellcolor[HTML]{EFEFEF}{\color[HTML]{00009B} \textbf{12.66}}} & {\color[HTML]{00009B}{\textbf{89.96}}} & \multicolumn{1}{c|}{\cellcolor[HTML]{EFEFEF}{\color[HTML]{00009B} \textbf{8.26}}}  & {\color[HTML]{00009B} {\textbf{91.66}}} & \multicolumn{1}{c|}{\cellcolor[HTML]{EFEFEF}{\color[HTML]{00009B} \textbf{2.86}}}          & {\color[HTML]{00009B}{\textbf{93.09}}} & \multicolumn{1}{c|}{\cellcolor[HTML]{EFEFEF}{\color[HTML]{00009B}\textbf{0.32}}}          & {\color[HTML]{00009B} \textbf{93.73}}                  \\
\rowcolor[HTML]{EFEFEF} 
\multicolumn{1}{l|}{\cellcolor[HTML]{EFEFEF}{\color[HTML]{000000} \textit{\textbf{Ours-2 (Generative Replay)}}}} & {\color[HTML]{9A0000} { \textbf{88.07}}} & \multicolumn{1}{c|}{\cellcolor[HTML]{EFEFEF}{\color[HTML]{000000} 60.86}}          & {\color[HTML]{9A0000} {\textbf{90.52}}} & \multicolumn{1}{c|}{\cellcolor[HTML]{EFEFEF}{\color[HTML]{9A0000} \textbf{5.20}}}  & {\color[HTML]{9A0000} {\textbf{91.44}}} & \multicolumn{1}{c|}{\cellcolor[HTML]{EFEFEF}{\color[HTML]{9A0000}\textbf{3.18}}}  & {\color[HTML]{9A0000} {\textbf{92.45}}} & \multicolumn{1}{c|}{\cellcolor[HTML]{EFEFEF}{\color[HTML]{9A0000}\textbf{1.45}}} & {\color[HTML]{9A0000} {\textbf{93.48}}} & \multicolumn{1}{c|}{\cellcolor[HTML]{EFEFEF}{\color[HTML]{000000} 0.18}}          & {\color[HTML]{9A0000}\textbf{94.02}}                  \\ \midrule
\multicolumn{12}{c}{\textbf{CIFAR100 \cite{CIFAR}}}                                                                                                                                                                                                                                                                                                                                                                                                                                                                                                                                                                                                                                                                                                                                                                                                                 \\
\multicolumn{12}{c}{$\boldsymbol{\mathcal{T}_{Acc} = 77.94\% }$, $\boldsymbol{\mathcal{S}_{Acc} = 77.10\%}$, \textbf{ ($\mathcal{E}_{max}$)}  = \textbf{400}}                                                                                                                                                                                                                                                                                                                                                                                                                                                                                                                                                                                                                                                                                                                                      \\ \midrule
\multicolumn{1}{l|}{{\color[HTML]{343434} DAFL \cite{DAFL}}}                                                                 & {\color[HTML]{343434} 52.48}                & \multicolumn{1}{c|}{{\color[HTML]{343434} 437.88}}                                 & {\color[HTML]{343434} 61.65}                & \multicolumn{1}{c|}{{\color[HTML]{343434} 82.96}}                                  & {\color[HTML]{343434} 65.88}                & \multicolumn{1}{c|}{{\color[HTML]{343434} 32.23}}                                  & {\color[HTML]{343434} 69.23}                & \multicolumn{1}{c|}{{\color[HTML]{343434} 11.93}}                                 & {\color[HTML]{343434} 72.23}                & \multicolumn{1}{c|}{{\color[HTML]{343434} 1.29}}                                  & {\color[HTML]{343434} 73.78}                           \\
\multicolumn{1}{l|}{{\color[HTML]{343434} DFAD \cite{fang2019data}}}                                                                 & {\color[HTML]{343434} 59.62}                & \multicolumn{1}{c|}{{\color[HTML]{343434} 192.64}}                                 & {\color[HTML]{343434} 65.32}                & \multicolumn{1}{c|}{{\color[HTML]{343434} 28.84}}                                  & {\color[HTML]{343434} 67.71}                & \multicolumn{1}{c|}{{\color[HTML]{343434} 2.72}}                                   & {\color[HTML]{343434} 68.69}                & \multicolumn{1}{c|}{{\color[HTML]{343434} 0.33}}                                  & {\color[HTML]{343434} 69.07}                & \multicolumn{1}{c|}{{\color[HTML]{343434} 0.24}}                                  & {\color[HTML]{343434} 69.73}                           \\
\multicolumn{1}{l|}{{\color[HTML]{343434} DFQ \cite{choi2020data}}}                                                                  & {\color[HTML]{343434} 68.20}                & \multicolumn{1}{c|}{{\color[HTML]{343434} 55.91}}                                  & {\color[HTML]{343434} 70.45}                & \multicolumn{1}{c|}{{\color[HTML]{343434} 9.78}}                                   & {\color[HTML]{343434} 71.77}                & \multicolumn{1}{c|}{{\color[HTML]{343434} 5.88}}                                   & {\color[HTML]{343434} 73.19}                & \multicolumn{1}{c|}{{\color[HTML]{343434} 2.22}}                                  & {\color[HTML]{343434} 74.48}                & \multicolumn{1}{c|}{{\color[HTML]{343434} 0.47}}                                  & {\color[HTML]{343434} 75.39}                           \\
\multicolumn{1}{l|}{{\color[HTML]{343434} DFKD\textsuperscript{*} \cite{binici2022preventing}}}                                                                 & {\color[HTML]{343434} 69.36}                & \multicolumn{1}{c|}{{\color[HTML]{343434} 74.67}}                                  & {\color[HTML]{343434} 72.35}                & \multicolumn{1}{c|}{{\color[HTML]{343434} 8.34}}                                   & {\color[HTML]{343434} 73.59}                & \multicolumn{1}{c|}{{\color[HTML]{343434} 4.54}}                                   & {\color[HTML]{343434} 74.86}                & \multicolumn{1}{c|}{{\color[HTML]{343434} 1.63}}                                  & {\color[HTML]{343434} 75.94}                & \multicolumn{1}{c|}{{\color[HTML]{343434} 0.18}}                                  & {\color[HTML]{343434} 76.51}                           \\ \midrule
\multicolumn{1}{l|}{{\color[HTML]{000000} MB-DFKD \cite{binici2022preventing}}}                                                              & {\color[HTML]{000000} 66.05}                & \multicolumn{1}{c|}{{\color[HTML]{000000} 207.29}}                                 & {\color[HTML]{000000} 71.16}                & \multicolumn{1}{c|}{{\color[HTML]{000000} 14.67}}                                  & {\color[HTML]{000000} 72.97}                & \multicolumn{1}{c|}{{\color[HTML]{000000} 5.61}}                                   & {\color[HTML]{000000} 74.40}                & \multicolumn{1}{c|}{{\color[HTML]{000000} 1.56}}                                  & {\color[HTML]{000000} 75.48}                & \multicolumn{1}{c|}{\textbf{\color[HTML]{00009B} 0.18}}                                  & {\color[HTML]{000000} 76.14}                           \\
\multicolumn{1}{l|}{{\color[HTML]{000000} PRE-DFKD \cite{binici2022robust}}}                                                             & {\color[HTML]{000000} 70.23}                & \multicolumn{1}{c|}{{\color[HTML]{000000} 86.63}}                                  & {\color[HTML]{000000} 73.39}                & \multicolumn{1}{c|}{{\color[HTML]{000000} 6.77}}                                   & {\color[HTML]{000000} 74.59}                & \multicolumn{1}{c|}{{\color[HTML]{000000} 2.88}}                                   & {\color[HTML]{000000} 75.60}                & \multicolumn{1}{c|}{{\color[HTML]{000000} 1.03}}                                  & {\color[HTML]{000000} 76.46}                & \multicolumn{1}{c|}{{\color[HTML]{000000} 0.12}}                                  & {\color[HTML]{000000} 76.93}                           \\
\rowcolor[HTML]{EFEFEF} 
\multicolumn{1}{l|}{\cellcolor[HTML]{EFEFEF}{\color[HTML]{000000} \textit{\textbf{Ours-1 (Memory Bank)}}}}       & {\color[HTML]{00009B} { \textbf{69.87}}} & \multicolumn{1}{c|}{\cellcolor[HTML]{EFEFEF}{\color[HTML]{00009B} \textbf{75.67}}} & {\color[HTML]{00009B} { \textbf{72.96}}} & \multicolumn{1}{c|}{\cellcolor[HTML]{EFEFEF}{\color[HTML]{00009B} \textbf{8.72}}}  & {\color[HTML]{00009B} { \textbf{74.30}}} & \multicolumn{1}{c|}{\cellcolor[HTML]{EFEFEF}{\color[HTML]{00009B} \textbf{4.03}}}  & {\color[HTML]{00009B} { \textbf{75.49}}} & \multicolumn{1}{c|}{\cellcolor[HTML]{EFEFEF}{\color[HTML]{00009B} \textbf{1.38}}} & {\color[HTML]{00009B} { \textbf{76.50}}} & \multicolumn{1}{c|}{\cellcolor[HTML]{EFEFEF}{\color[HTML]{000000} 0.20}}          & {\color[HTML]{00009B} \textbf{77.11}}                  \\
\rowcolor[HTML]{EFEFEF} 
\multicolumn{1}{l|}{\cellcolor[HTML]{EFEFEF}{\color[HTML]{000000} \textit{\textbf{Ours-2 (Generative Replay)}}}} & {\color[HTML]{9A0000} { \textbf{71.49}}} & \multicolumn{1}{c|}{\cellcolor[HTML]{EFEFEF}{\color[HTML]{9A0000} \textbf{60.17}}} & {\color[HTML]{9A0000} { \textbf{74.16}}} & \multicolumn{1}{c|}{\cellcolor[HTML]{EFEFEF}{\color[HTML]{9A0000} \textbf{4.61}}}  & {\color[HTML]{9A0000} { \textbf{75.12}}} & \multicolumn{1}{c|}{\cellcolor[HTML]{EFEFEF}{\color[HTML]{9A0000} \textbf{2.26}}}  & {\color[HTML]{9A0000} { \textbf{76.01}}} & \multicolumn{1}{c|}{\cellcolor[HTML]{EFEFEF}{\color[HTML]{9A0000} \textbf{0.74}}} & {\color[HTML]{9A0000} { \textbf{76.75}}} & \multicolumn{1}{c|}{\cellcolor[HTML]{EFEFEF}{\color[HTML]{9A0000} \textbf{0.09}}} & {\color[HTML]{9A0000} \textbf{77.21}}                  \\ \midrule
\multicolumn{12}{c}{\textbf{Tiny-ImageNet \cite{tiny}}}                                                                                                                                                                                                                                                                                                                                                                                                                                                                                                                                                                                                                                                                                                                                                                                                            \\
\multicolumn{12}{c}{$\boldsymbol{\mathcal{T}_{Acc} = 60.83 \%}$, $\boldsymbol{\mathcal{S}_{Acc} = 57.88\%}$, \textbf{ ($\mathcal{E}_{max}$)}  = \textbf{500}}                                                                                                                                                                                                                                                                                                                                                                                                                                                                                                                                                                                                                                                                                                                                      \\ \midrule
\multicolumn{1}{l|}{{\color[HTML]{343434} DAFL \cite{DAFL}}}                                                                 & {\color[HTML]{343434} 21.04}                & \multicolumn{1}{c|}{{\color[HTML]{343434} 106.04}}                                 & {\color[HTML]{343434} 24.95}                & \multicolumn{1}{c|}{{\color[HTML]{343434} 52.56}}                                  & {\color[HTML]{343434} 28.10}                & \multicolumn{1}{c|}{{\color[HTML]{343434} 28.22}}                                  & {\color[HTML]{343434} 31.19}                & \multicolumn{1}{c|}{{\color[HTML]{343434} 11.44}}                                 & {\color[HTML]{343434} 34.11}                & \multicolumn{1}{c|}{{\color[HTML]{343434} 1.33}}                                  & {\color[HTML]{343434} 35.46}                           \\
\multicolumn{1}{l|}{{\color[HTML]{343434} DFAD \cite{fang2019data}}}                                                                 & {\color[HTML]{343434} 14.60}                & \multicolumn{1}{c|}{{\color[HTML]{343434} 22.32}}                                  & {\color[HTML]{343434} 16.48}                & \multicolumn{1}{c|}{{\color[HTML]{343434} 7.48}}                                   & {\color[HTML]{343434} 17.76}                & \multicolumn{1}{c|}{{\color[HTML]{343434} 1.43}}                                   & {\color[HTML]{343434} 18.44}                & \multicolumn{1}{c|}{{\color[HTML]{343434} 0.26}}                                  & {\color[HTML]{343434} 18.84}                & \multicolumn{1}{c|}{{\color[HTML]{343434} 0.10}}                                  & {\color[HTML]{343434} 19.60}                           \\
\multicolumn{1}{l|}{{\color[HTML]{343434} DFKD\textsuperscript{*} \cite{binici2022preventing}}}                                                                 & {\color[HTML]{343434} 34.55}                & \multicolumn{1}{c|}{{\color[HTML]{343434} 86.20}}                                  & {\color[HTML]{343434} 38.17}                & \multicolumn{1}{c|}{{\color[HTML]{343434} 23.88}}                                  & {\color[HTML]{343434} 40.28}                & \multicolumn{1}{c|}{{\color[HTML]{343434} 13.28}}                                  & {\color[HTML]{343434} 42.40}                & \multicolumn{1}{c|}{{\color[HTML]{343434} 5.32}}                                  & {\color[HTML]{343434} 44.38}                & \multicolumn{1}{c|}{{\color[HTML]{343434} 0.79}}                                  & {\color[HTML]{343434} 45.61}                           \\ \midrule
\multicolumn{1}{l|}{{\color[HTML]{000000} MB-DFKD \cite{binici2022preventing}}}                                                              & {\color[HTML]{000000} 34.16}                & \multicolumn{1}{c|}{{\color[HTML]{000000} 122.92}}                                 & {\color[HTML]{000000} 38.77}                & \multicolumn{1}{c|}{{\color[HTML]{000000} 32.42}}                                  & {\color[HTML]{000000} 41.18}                & \multicolumn{1}{c|}{{\color[HTML]{000000} 18.40}}                                  & {\color[HTML]{000000} 43.63}                & \multicolumn{1}{c|}{{\color[HTML]{000000} 8.48}}                                  & {\color[HTML]{000000} 46.17}                & \multicolumn{1}{c|}{{\color[HTML]{000000} 1.73}}                                  & {\color[HTML]{000000} 47.96}                           \\
\multicolumn{1}{l|}{{\color[HTML]{000000} PRE-DFKD \cite{binici2022robust}}}                                                             & {\color[HTML]{000000} 38.89}                & \multicolumn{1}{c|}{{\color[HTML]{000000} 80.12}}                                  & {\color[HTML]{000000} 42.27}                & \multicolumn{1}{c|}{{\color[HTML]{000000} 21.66}}                                  & {\color[HTML]{9A0000}\textbf{44.25}}                & \multicolumn{1}{c|}{{\color[HTML]{9A0000}\textbf{12.61}}}                                  & {\color[HTML]{9A0000}\textbf{46.31}}                & \multicolumn{1}{c|}{{\color[HTML]{9A0000}\textbf{5.47}}}                                  & {\color[HTML]{000000} 48.33}                & \multicolumn{1}{c|}{{\color[HTML]{000000} 1.39}}                                  & {\color[HTML]{9A0000}\textbf{49.94}}                           \\
\rowcolor[HTML]{EFEFEF} 
\multicolumn{1}{l|}{\cellcolor[HTML]{EFEFEF}{\color[HTML]{000000}\textit{\textbf{Ours-1 (Memory Bank)}}}}       & {\color[HTML]{00009B} {\textbf{36.34}}} & \multicolumn{1}{c|}{\cellcolor[HTML]{EFEFEF}{\color[HTML]{00009B} \textbf{94.62}}} & {\color[HTML]{00009B} {\textbf{40.03}}} & \multicolumn{1}{c|}{\cellcolor[HTML]{EFEFEF}{\color[HTML]{00009B}\textbf{26.20}}} & {\color[HTML]{00009B} {\textbf{42.28}}} & \multicolumn{1}{c|}{\cellcolor[HTML]{EFEFEF}{\color[HTML]{00009B}\textbf{13.76}}} & {\color[HTML]{00009B} {\textbf{44.43}}} & \multicolumn{1}{c|}{\cellcolor[HTML]{EFEFEF}{\color[HTML]{00009B} \textbf{5.82}}} & {\color[HTML]{00009B} {\textbf{46.52}}} & \multicolumn{1}{c|}{\cellcolor[HTML]{EFEFEF}{\color[HTML]{00009B}\textbf{0.95}}} & {\color[HTML]{00009B} \textbf{47.96}}                           \\
\rowcolor[HTML]{EFEFEF} 
\multicolumn{1}{l|}{\cellcolor[HTML]{EFEFEF}{\color[HTML]{000000} \textit{\textbf{Ours-2 (Generative Replay)}}}} & {\color[HTML]{9A0000} {\textbf{39.09}}} & \multicolumn{1}{c|}{\cellcolor[HTML]{EFEFEF}{\color[HTML]{9A0000} \textbf{74.95}}} & {\color[HTML]{9A0000} { \textbf{42.30}}} & \multicolumn{1}{c|}{\cellcolor[HTML]{EFEFEF}{\color[HTML]{9A0000} \textbf{21.23}}} & {\color[HTML]{000000} 44.19}                & \multicolumn{1}{c|}{\cellcolor[HTML]{EFEFEF}{\color[HTML]{000000} 13.55}}          & {\color[HTML]{000000} 46.30}                & \multicolumn{1}{c|}{\cellcolor[HTML]{EFEFEF}{\color[HTML]{000000} 6.21}}          & {\color[HTML]{9A0000} { \textbf{48.48}}} & \multicolumn{1}{c|}{\cellcolor[HTML]{EFEFEF}\textbf{{\color[HTML]{9A0000} 1.12}}} & {\color[HTML]{000000} 49.88}                           \\ \bottomrule
\end{tabular}%
}

\label{tab:main_table}
\end{table*}
\subsection{Generator Update in Adversarial Exploration-based DFKD}
\label{sec:generator}
In the absence of the training dataset ($\mathcal{D}_{\mathcal{T}}$), the generator ($\mathcal{G}$) is utilized to obtain pseudo-samples ($\hat{x}$) and perform knowledge-distillation, \textit{i.e.}, $\hat{x} = \mathcal{G}(z)$, $z \sim \mathcal{N}(0,I)$. The generator is learned to maximize the disagreement between Teacher network ($\mathcal{T}_{\theta_{\mathcal{T}}}$) and the Student network ($\mathcal{S}_{\theta_{\mathcal{S}}}$). Additionally, for the generated data $\hat{x}$ to mimic similar responses from the teacher as the real data, we include a prior loss $\mathcal{L}_{\mathcal{P}}$ \cite{DAFL} to be minimized alongside maximizing the discrepancy ($D$). Hence, we update the generator parameters ($\theta_{\mathcal{G}}$) by maximizing the following objective:
\begin{align}
\max_{\theta_{\mathcal{G}}} \mathbb{E}_{z\sim\mathcal{N}(0,I)}[D(\mathcal{T}_{\theta_{\mathcal{T}}}(\mathcal{G}_{\theta_{\mathcal{G}}}(z)), & \mathcal{S}_{\theta_{\mathcal{S}}}(\mathcal{G}_{\theta_{\mathcal{G}}}(z))) \nonumber\\  &- \mathcal{L}_{\mathcal{P}}(\mathcal{G}_{\theta_{\mathcal{G}}}(z))].
\end{align}
Typically the disagreement function ($D$) for the generator is identical to the teacher-student disagreement term \cite{fang2019data,choi2020data}. Instead, for teacher-student discrepancy maximization we use the Jensen-Shannon (JS) ($\mathcal{L}_{JS}$) divergence. Our motivation to use JS divergence is based on empirical study by Binici \textit{et al.}~\cite{binici2022preventing}.  Hence, $D$ is defined as:
\begin{align}
     D&(a, b) = \mathcal{L}_{JS}(p(a),p(b)), \nonumber \\
     \mathcal{L}_{JS}(p(a),p(b)) = &\frac{1}{2}(\mathcal{L}_{KL}(p(a),m)+\mathcal{L}_{KL}(p(b),m)), \text{and}\nonumber \\ 
    &m = \frac{1}{2}(p(a)+p(b)).
\end{align}
Here $\mathcal{L}_{KL}$ stands for the  Kullback–Leibler divergence and $p(a)$ and $p(b)$ denote the probability vectors obtained after the $\softmax$ applied to the arguments $a$ and $b$, respectively.

Moreover, the prior loss $\mathcal{L_{P}}$ \cite{binici2022preventing} is defined as the combination of three loss functions ($\mathcal{L}_{OH}$, $\mathcal{L}_{A}$, and $\mathcal{L}_{EM}$) that capture different characteristics from the teacher model and impose them on the pseudo samples $\hat{x}$, and is defined as:
\begin{equation}
    \mathcal{L_{P}} = \mathcal{L}_{OH} + \gamma \mathcal{L}_{A} + \delta\mathcal{L}_{EM},
\end{equation}
where, $\gamma$ and $\delta$ denote the weighing scalar coefficients.
\noindent \textbullet \ $\mathcal{L}_{OH}$ is the one-hot loss that forces the generated samples to have strong (one-hot vector like) predictions when input to the teacher. It is defined as the cross-entropy between the teacher's softmax output $p_{\mathcal{T}}(\hat{x}_{n}) = \softmax(\mathcal{T}_{\theta_{\mathcal{T}}}(\hat{x}_{n})), \hat{x}_{n} \in \hat{x}$, and its one-hot vector version $e_{c} \in \{0,1\}^{C}$, where $C$ denotes the total number of classes and $e_{c}$ denotes the $c$-th canonical one-hot-vector, and $c = \argmax_{i}(p_{\mathcal{T}}^{i}(\hat{x}_{n}))$, the superscript $i$ denotes the $i^{th}$ probability score of the predicted mass vector $p_{\mathcal{T}}(\hat{x}_{n})$. Hence, $\mathcal{L}_{OH}$ is defined as:
\begin{equation}
    \mathcal{L}_{OH} = -\frac{1}{N}\sum_{n=1}^{N}e_{c}^{\top}\log(p_{\mathcal{T}}(\hat{x}_{n})).
\end{equation} 
\noindent \textbullet \ $\mathcal{L}_{A}$ is the activation loss motivated by the notion that meaningful inputs result in higher-valued activation maps in a well-trained network \cite{DAFL} and is defined as:
\begin{equation}
    \mathcal{L}_{A} = -\frac{1}{NL}\sum_{n=1}^{N}\sum_{l=1}^{L}{\lVert \mathcal{A}_{\mathcal{T}}^{l}(\hat{x}_{n}) \rVert_{1}},
\end{equation}
where $\mathcal{A}_{\mathcal{T}}^{l}(\hat{x}_{n})$ denotes the activation of the $l$-th layer in the Teacher network ($\mathcal{T}_{\theta_{\mathcal{T}}}$) for the input $n^{th}$ input $\hat{x}_{n} \in \hat{x}$. 

\noindent \textbullet \ $\mathcal{L}_{EM}$ loss is the entropy-maximization term imposed on the generator to output an equal number of pseudo-samples from each category \cite{choi2020data}. In other words, the intra-batch entropy is maximized, resulting in a similar number of samples for each category \textit{i.e.} if $\bar{p}_{\mathcal{T}} = \frac{1}{N}\sum_{n=1}^{N}p_{\mathcal{T}}(\hat{x}_{n})$, then the loss $\mathcal{L}_{EM}$ is defined as:
\begin{equation}
    \mathcal{L}_{EM} = \bar{p}_{\mathcal{T}}^{\top}\log(\bar{p}_{\mathcal{T}}).
\end{equation}
In sum, the Generator loss objective ($\mathcal{L}_{\mathcal{G}}$) is defined as:
\begin{equation}
    \mathcal{L_{G}}(\theta_{\mathcal{G}}) = -D(\mathcal{T}_{\theta_{\mathcal{T}}}(\mathcal{G}_{\theta_{\mathcal{G}}}(z)), \mathcal{S}_{\theta_{\mathcal{S}}}(\mathcal{G}_{\theta_{\mathcal{G}}}(z))) + \mathcal{L}_{\mathcal{P}}(\mathcal{G}_{\theta_{\mathcal{G}}}(z)).
\end{equation}
\begin{figure*}[t]
    \centering
    \includegraphics[width=\textwidth]{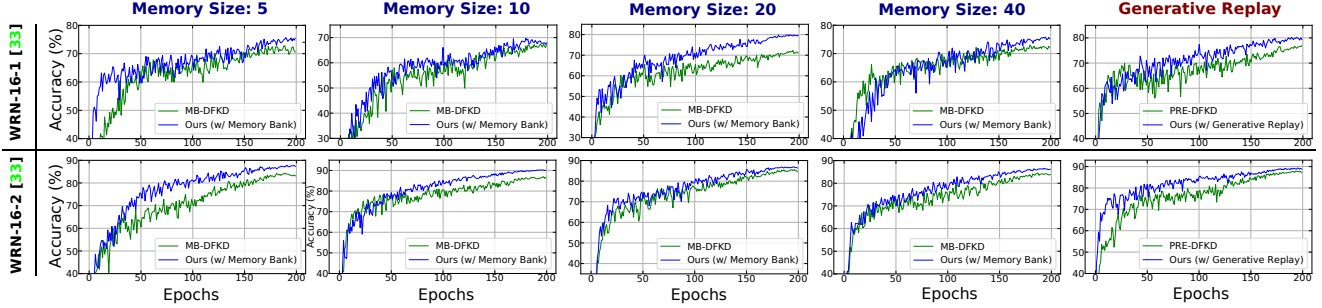}
    \caption{Student learning curves depicting the learning evolution of Wide-ResNet (WRN) \cite{WRN}. The WRN-16-1 (\textit{top-row}), and WRN-16-2 (\textit{bottom-row}) networks are distilled by a WRN-40-2 teacher network pre-trained on CIFAR10 ($\mathcal{T}_{Acc} = 94.87\%$). Each column represent the learning curves with the Buffer-based (with different memory buffer sizes) and Generative replay schemes. The proposed method is in \textbf{\color[HTML]{3b80ee} Blue}.}
    \label{fig:wrn_cifar10}
\end{figure*}
\section{Experiments}
\subsection{Experimental Settings}
\label{sec:experiments}
\noindent \textbf{Datasets:} We evaluate the proposed method on SVHN \cite{SVHN}, CIFAR10 \cite{CIFAR}, CIFAR100 \cite{CIFAR} and Tiny-ImageNet \cite{tiny} datasets.\\\textbf{Teacher model: }For CIFAR10 and CIFAR100 datasets, we used the pre-trained teacher models made available by the authors of \cite{fang2021contrastive} and \cite{binici2022preventing}, respectively. For SVHN and Tiny-ImageNet we trained the teacher network from scratch. We provide the training details of the teacher models in the Supplemental Material.\\\textbf{Definition of an Epoch in DFKD: }In Adversarial DFKD, the notion of an \emph{epoch} is obscure. In a typical deep neural network-based classification training, an epoch is defined as one complete pass over the available training data. However, DFKD has no access to the training data; instead, the pseudo samples generated on-the-fly are used to distill knowledge to the student network. Therefore, prior works \cite{fang2019data,choi2020data,binici2022preventing} defined an epoch in terms of a fixed number of training iterations ($\mathcal{I}$), where each iteration consists of a set number of generator update steps ($g$) and student update steps ($s$). Hence, to be consistent across the baselines and prior arts, we use the same number of training iterations, generator update steps, and student updates steps to define an epoch. For all the methods, we set $\mathcal{I} = 72$, $g = 1$, and $s = 10$ and use a batch size of $512$ of the sampled noise ($z$) to generate the pseudo samples and optimize the parameters $\theta_{\mathcal{G}}$ and $\theta_{\mathcal{S}}$.\\\textbf{Training Details: }Due to the page-limit constraint, the training details are provided in the Supplemental Material.\\\textbf{Evaluation: }We evaluate the methods by comparing the mean and variance of the student network's test accuracy ($\mathcal{S}_{Acc}$), denoting them as $\mu[\mathcal{S}_{Acc}]$, and $\sigma^{2}[\mathcal{S}_{Acc}]$, respectively, across the epochs, motivated by Binci \textit{et al.} \cite{binici2022robust}. Specifically, we compare the different sections of the student's learning evolution by partitioning them into different epoch percentiles. For example, computing the $\mu[\mathcal{S}_{Acc}]$ and $\sigma^{2}[\mathcal{S}_{Acc}]$ for epochs greater than the $n^{th}$ percentile conveys the mean and variance across all the epochs greater than the ${\frac{n}{100}}^{th}$ of the total number of training epochs.
\subsection{Results and Observations}
\label{sec:mu_sigma}
\noindent \textbf{Baseline and State-of-the-art Comparisons}
In Table \ref{tab:main_table}, we analyze our method on classification task and compare it with prior Adversarial DFKD methods \cite{DAFL,fang2019data,choi2020data} and closely related memory-based Adversarial DFKD methods \cite{binici2022preventing,binici2022robust}. For a fair comparison across the methods, we re-implemented all the methods and used the same generator architecture to generate the pseudo samples. For each of the methods we report the $\mu[\mathcal{S}_{Acc}]$ and $\sigma^{2}[{\mathcal{S}_{Acc}}]$ at different epoch percentiles and the maximum accuracy (Acc\textsubscript{max} (\%)) attained by the student. We observe that, compared to MB-DFKD (Memory Bank) \cite{binici2022preventing}, \emph{Ours-1} (Memory Bank) demonstrates consistent improvement across all the datasets. Similarly, compared to PRE-DFKD (Generative Replay) \cite{binici2022robust}, utilizing the same VAE decoder architecture as the generative replay, we observe a similar trend for \emph{Ours-2} (Generative Replay).
\begin{table*}[ht]
\centering
\resizebox{\textwidth}{!}{%
\begin{tabular}{@{}lccccccccccccc
>{\columncolor[HTML]{EFEFEF}}c 
>{\columncolor[HTML]{EFEFEF}}c @{}}
\toprule
                                                                       & \textbf{ZSKD\textsuperscript{a} \cite{nayak2019zero}} & \textbf{ADI\textsuperscript{b} \cite{yin2020dreaming}} & \textbf{CMI\textsuperscript{b} \cite{fang2021contrastive}}               & \textbf{DeGAN\textsuperscript{c} \cite{DeGAN}} & \textbf{EATSKD\textsuperscript{c} \cite{EATSKD}}            & \textbf{KEGNET\textsuperscript{a}\cite{KEGNET}} & \textbf{ZSKT\textsuperscript{b} \cite{micaelli2019zero}} & \textbf{DDAD\textsuperscript{a} \cite{DDAD}} & \textbf{DAFL\textsuperscript{d} \cite{DAFL}} & \textbf{DFAD\textsuperscript{d} \cite{fang2019data}} & \textbf{DFQ\textsuperscript{d} \cite{choi2020data}}               & \textbf{MB-DFKD\textsuperscript{d} \cite{binici2022preventing}} & \textbf{PRE-DFKD\textsuperscript{d} \cite{binici2022robust}} & \textit{\textbf{Ours-1}}                       & \textit{\textbf{Ours-2}}  \\ \midrule
\multicolumn{1}{l|}{{\color[HTML]{656565} $\mathcal{T}_{Acc}$ (\%)}}  & {\color[HTML]{656565} 77.50} & {\color[HTML]{656565} 78.05} & \multicolumn{1}{c|}{{\color[HTML]{656565} 78.05}} & {\color[HTML]{656565} 77.94} & \multicolumn{1}{c|}{{\color[HTML]{656565} 77.94}} & {\color[HTML]{656565} 77.50} & {\color[HTML]{656565} 78.05} & {\color[HTML]{656565} 77.50} & {\color[HTML]{656565} 77.94} & {\color[HTML]{656565} 77.94} & \multicolumn{1}{c|}{{\color[HTML]{656565} 77.94}} & {\color[HTML]{656565} 77.94} & {\color[HTML]{656565} 77.94} & {\color[HTML]{656565} 77.94}          & {\color[HTML]{656565} 77.94}          \\
\multicolumn{1}{l|}{$\mathcal{S}_{Acc}$ (\%)}                         & 70.21                        & 61.32                        & \multicolumn{1}{c|}{77.04}                        & 65.25                        & \multicolumn{1}{c|}{67.18}                        & 73.91                        & 67.74                        & 75.04                        & 73.79                        & 69.73                        & \multicolumn{1}{c|}{75.39}                        & 76.14                        & 76.93                        & {\color[HTML]{00009B} \textbf{77.11}} & {\color[HTML]{9A0000} \textbf{77.21}} \\
\multicolumn{1}{l|}{$\Delta_{\mathcal{T}_{Acc} - \mathcal{S}_{Acc}}$} & 7.29                         & 16.73                        & \multicolumn{1}{c|}{1.01}                         & 12.69                        & \multicolumn{1}{c|}{10.76}                        & 3.59                         & 10.31                        & 2.46                         & 4.15                         & 8.21                         & \multicolumn{1}{c|}{2.55}                         & 1.80                         & 1.01                         & {\color[HTML]{00009B} \textbf{0.83}}  & {\color[HTML]{9A0000} \textbf{0.73}}  \\ \bottomrule
\end{tabular}%
}
\caption{Classification accuracy (in \%) of the student trained using various DFKD methods on CIFAR100 with ResNet-34 \cite{he2016deep} as the teacher and ResNet-18 \cite{he2016deep} as the student. $\mathcal{T}_{Acc}$ and  $\mathcal{S}_{Acc}$ denote the Teacher network's and the Student network's accuracy, respectively, and $\Delta_{\mathcal{T}_{Acc} - \mathcal{S}_{Acc}}$ denotes the difference between the Teacher and Student accuracies. Also, \textsuperscript{a}, \textsuperscript{b}, \textsuperscript{c} and \textsuperscript{d} denote results produced by \cite{DDAD}, \cite{fang2021contrastive}, \cite{binici2022preventing}, and our implementation, respectively.}\label{tab:cifar100-comparison}
\end{table*}

Moreover, in Figure \ref{fig:learning_curves}, we visualize the learning curves of the student networks trained on multiple datasets. We plot the test accuracy of the student at the end of each training epoch. The proposed method exhibits significant improvement in the learning evolution and the peak accuracies achieved, suggesting that the proposed approach can retain the knowledge from previously encountered samples as the learning progresses. However, on Tiny-ImageNet, with Generative replay, we did not observe substantial improvements; we conjecture that this may be due to the complexity of the dataset, and the inability to capture crucial samples as replay for the complex dataset, for a large number of epochs. Also, with Generative Replay we sometimes faced difficulty in training a VAE on a stream of synthetic samples (especially for complex dataset like Tiny-ImageNet) as it suffers due to the distribution drift of its own. 

Additionally, we emphasize that we do not strive toward achieving state-of-the-art student classification accuracy (requiring extensive hyper-parameter tuning) in the DFKD setting, but verify the viability of our hypothesis of retaining the previously acquired knowledge while learning on new samples. Nonetheless, we observe that our method improves upon the student classification accuracy on CIFAR100 \cite{CIFAR} compared to the contemporary works and the current state-of-the-art \cite{binici2022robust} with the ResNet-34 \cite{he2016deep} ($\mathcal{T}$) and ResNet-18 \cite{he2016deep} ($\mathcal{S}$) setup, as shown in Table \ref{tab:cifar100-comparison}. Additionally, since previous works use the same teacher network with different test accuracies, we also report the teacher accuracies of the respective methods used to distill the knowledge to the student. Nonetheless, we also compute the Teacher-Student accuracy difference ($\Delta_{\mathcal{T}_{Acc} - \mathcal{S}_{Acc}}$) to assess the distance of the student from its teacher in terms of classification accuracy.\\
\noindent \textbf{Interpreting the Result: } Because of the proposed student update strategy, we observe a global monotonicity in the student's learning evolution which the existing approaches with naive replay \cite{binici2022preventing,binici2022robust} claim, but do not warrant (described in Section \ref{sec:LRA}). The global monotonicity in the learning evolution encompasses crucial advantages. For example, when the validation data is unavailable, the developer cannot assess the student's accuracy and identify the best parameters for the student. In such a scenario, the final accuracy is dependent on the random termination epoch set by the developer. In other words, the ideal DFKD approach should sustain high accuracy via monotonically enhancing it during the course of distillation. Therefore, $\mu[\mathcal{S}_{Acc}]$ and $\sigma^{2}[\mathcal{S}_{Acc}]$ contribute as crucial metrics to asses the distillation method as opposed to the maximum accuracy (Acc\textsubscript{max}), since the Acc\textsubscript{max} value can be attained at any point of time prior to the termination, and can be misleading. The improvements in the monotonicity and the $\mu[S_{Acc}]$ and $\sigma^{2}[S_{Acc}]$ values of proposed method are evident from Table \ref{tab:main_table}, Figure \ref{fig:learning_curves} and Figure \ref{fig:wrn_cifar10}.\\
\noindent \textbf{Architecture Scalability:} The proposed student update strategy is generic and is scalable across different neural network architecture families since the method is not constrained to a particular choice of architecture. From the  ResNet-18~\cite{he2016deep}, WRN-16-1~\cite{WRN} and WRN-16-2~\cite{WRN} student learning curves~(in Figure~\ref{fig:learning_curves} and Figure~\ref{fig:wrn_cifar10}), we observe our method’s advantage on both the network architectures. Moreover, for large student network architectures (Deeper or Wider) that include complex layers, the proposed method efficiently handles the intricacies with regard to computing the Hessian and the Hessian-product, which becomes highly essential for cumbersome models.\\\textbf{Improvement across Replay Schemes:} Furthermore, the proposed method is agnostic to the memory scheme employed for replay, as demonstrated by the experiments (in Table~\ref{tab:main_table}, Figure~\ref{fig:learning_curves} and Figure~\ref{fig:wrn_cifar10}) using a Memory Buffer and Generative-Replay, thus, rendering our method generalized to the choice of replay. In  Table~\ref{tab:main_table} and Figure \ref{fig:wrn_cifar10}, we can observe that the proposed method enhances the student's performance on both the replay schemes (Memory Bank and Generative Replay) used in the prior arts. Moreover, we experiment with different memory buffer sizes on WRN-16-1~\cite{WRN} and WRN-16-2~\cite{WRN} distillation (in Figure~\ref{fig:wrn_cifar10}) and observe consistent and substantial improvements across different memory sizes. Here, the memory size is defined as the maximum number of pseudo-example batches that the bank can contain and each batch consists of randomly sampled 64 examples from $\hat{x}$.\\\textbf{GPU Memory Utilization:} Moreover, our student update strategy brings in no practical memory overhead, compared to memory-based Adversarial DFKD methods. We observe only a minimal increase in the GPU memory usage of few MBs ($\approx$ 40 MB) due to the higher order gradients computed as a part of the update on $\theta_{\mathcal{S}}$ through $\theta_{\mathcal{S}}^{\prime}$. Moreover, we use a \emph{single} gradient descent step to obtain $\theta_{\mathcal{S}}^{\prime}$, which does not incur a large memory overhead. Thus, we do not opt for a first order approximation~\cite{nichol2018first} of our method, which is much prevalent in the meta-learning literature.
\section{Conclusion}
\label{sec:conclusion}
\noindent \textbf{Societal Impact: } Similar to other DFKD methods, our method may be framed as an attack strategy to create clones of proprietary pre-trained models that are accessible online \cite{DFME}. However, this work makes no such efforts and does not support such practices.\\\textbf{Summary: }In this paper, we proposed a meta-learning inspired student update strategy for the Adversarial DFKD setting, that treats \emph{Knowledge-Acquisition} and \emph{Knowledge-Retention} as meta-train and meta-test, respectively. The proposed strategy substantially improves the learning evolution of the student network by implicitly aligning the $\emph{Knowledge-Retention}$ and the $\emph{Knowledge-Acquisition}$ tasks. The intended effect of having the gradient directions aligned is to obtain student parameters ($\theta_{\mathcal{S}}$) that have optimal performance on both $\mathcal{L}_{Acq}$ and $\mathcal{L}_{Ret}$. The conducted experiments on multiple datasets, network architectures, and replay schemes demonstrate the effectiveness, scalability and generalizability of the proposed strategy.

{\small
\bibliographystyle{ieee_fullname}
\bibliography{11_references}

\begin{thebibliography}{10}\itemsep=-1pt

\bibitem{DeGAN}
Sravanti Addepalli, Gaurav~Kumar Nayak, Anirban Chakraborty, and Venkatesh~Babu
  Radhakrishnan.
\newblock Degan: Data-enriching gan for retrieving representative samples from
  a trained classifier.
\newblock In {\em AAAI}, 2020.

\bibitem{binici2022robust}
Kuluhan Binici, Shivam Aggarwal, Nam~Trung Pham, Karianto Leman, and Tulika
  Mitra.
\newblock Robust and resource-efficient data-free knowledge distillation by
  generative pseudo replay.
\newblock In {\em AAAI}, 2022.

\bibitem{binici2022preventing}
Kuluhan Binici, Nam~Trung Pham, Tulika Mitra, and Karianto Leman.
\newblock Preventing catastrophic forgetting and distribution mismatch in
  knowledge distillation via synthetic data.
\newblock In {\em WACV}, 2022.

\bibitem{DAFL}
Hanting Chen, Yunhe Wang, Chang Xu, Zhaohui Yang, Chuanjian Liu, Boxin Shi,
  Chunjing Xu, Chao Xu, and Qi Tian.
\newblock Dafl: Data-free learning of student networks.
\newblock In {\em ICCV}, 2019.

\bibitem{choi2020data}
Yoojin Choi, Jihwan Choi, Mostafa El-Khamy, and Jungwon Lee.
\newblock Data-free network quantization with adversarial knowledge
  distillation.
\newblock In {\em CVPR Workshops}, 2020.

\bibitem{dandi2022implicit}
Yatin Dandi, Luis Barba, and Martin Jaggi.
\newblock Implicit gradient alignment in distributed and federated learning.
\newblock In {\em AAAI}, 2022.

\bibitem{antiforgetting2}
Matthias De~Lange, Rahaf Aljundi, Marc Masana, Sarah Parisot, Xu Jia,
  Ale{\v{s}} Leonardis, Gregory Slabaugh, and Tinne Tuytelaars.
\newblock A continual learning survey: Defying forgetting in classification
  tasks.
\newblock {\em IEEE TPAMI}, 44(7):3366--3385, 2021.

\bibitem{do2022momentum}
Kien Do, Hung Le, Dung Nguyen, Dang Nguyen, Haripriya Harikumar, Truyen Tran,
  Santu Rana, and Svetha Venkatesh.
\newblock Momentum adversarial distillation: Handling large distribution shifts
  in data-free knowledge distillation.
\newblock In {\em NeurIPS}, 2022.

\bibitem{fang2022up}
Gongfan Fang, Kanya Mo, Xinchao Wang, Jie Song, Shitao Bei, Haofei Zhang, and
  Mingli Song.
\newblock Up to 100x faster data-free knowledge distillation.
\newblock In {\em AAAI}, 2022.

\bibitem{fang2019data}
Gongfan Fang, Jie Song, Chengchao Shen, Xinchao Wang, Da Chen, and Mingli Song.
\newblock Data-free adversarial distillation.
\newblock In {\em CVPR}, 2020.

\bibitem{fang2021contrastive}
Gongfan Fang, Jie Song, Xinchao Wang, Chen Shen, Xingen Wang, and Mingli Song.
\newblock Contrastive model inversion for data-free knowledge distillation.
\newblock In {\em IJCAI}, 2021.

\bibitem{finn2017model}
Chelsea Finn, Pieter Abbeel, and Sergey Levine.
\newblock Model-agnostic meta-learning for fast adaptation of deep networks.
\newblock In {\em ICML}, 2017.

\bibitem{GAN}
Ian~J. Goodfellow, Jean Pouget-Abadie, Mehdi Mirza, Bing Xu, David
  Warde-Farley, Sherjil Ozair, Aaron~C. Courville, and Yoshua Bengio.
\newblock Generative adversarial nets.
\newblock In {\em NeurIPS}, 2014.

\bibitem{he2016deep}
Kaiming He, Xiangyu Zhang, Shaoqing Ren, and Jian Sun.
\newblock Deep residual learning for image recognition.
\newblock In {\em CVPR}, 2016.

\bibitem{VAE}
Diederik~P Kingma, Max Welling, et~al.
\newblock An introduction to variational autoencoders.
\newblock {\em Foundations and Trends{\textregistered} in Machine Learning},
  12(4):307--392, 2019.

\bibitem{CIFAR}
Alex Krizhevsky et~al.
\newblock Learning multiple layers of features from tiny images.
\newblock 2009.

\bibitem{tiny}
Ya Le and Xuan~S. Yang.
\newblock Tiny imagenet visual recognition challenge.
\newblock 2015.

\bibitem{CuDFKD}
Jingru Li, Sheng Zhou, Liangcheng Li, Xifeng Yan, Zhi Yu, and Jiajun Bu.
\newblock How to teach: Learning data-free knowledge distillation from
  curriculum.
\newblock {\em arXiv preprint arXiv:2208.13648}, 2022.

\bibitem{liu2021zero}
Yuang Liu, Wei Zhang, and Jun Wang.
\newblock Zero-shot adversarial quantization.
\newblock In {\em CVPR}, 2021.

\bibitem{micaelli2019zero}
Paul Micaelli and Amos~J Storkey.
\newblock Zero-shot knowledge transfer via adversarial belief matching.
\newblock In {\em NeurIPS}, 2019.

\bibitem{EATSKD}
Gaurav~Kumar Nayak, Konda~Reddy Mopuri, and Anirban Chakraborty.
\newblock Effectiveness of arbitrary transfer sets for data-free knowledge
  distillation.
\newblock In {\em WACV}, 2021.

\bibitem{nayak2019zero}
Gaurav~Kumar Nayak, Konda~Reddy Mopuri, Vaisakh Shaj, Venkatesh~Babu
  Radhakrishnan, and Anirban Chakraborty.
\newblock Zero-shot knowledge distillation in deep networks.
\newblock In {\em ICML}, 2019.

\bibitem{SVHN}
Yuval Netzer, Tao Wang, Adam Coates, Alessandro Bissacco, Bo Wu, and Andrew~Y.
  Ng.
\newblock Reading digits in natural images with unsupervised feature learning.
\newblock In {\em NeurIPS Workshop on Deep Learning and Unsupervised Feature
  Learning}, 2011.

\bibitem{nichol2018first}
Alex Nichol, Joshua Achiam, and John Schulman.
\newblock On first-order meta-learning algorithms.
\newblock {\em arXiv preprint arXiv:1803.02999}, 2018.

\bibitem{ZeroGrad}
Mert~Bulent Sariyildiz and Ramazan~Gokberk Cinbis.
\newblock Gradient matching generative networks for zero-shot learning.
\newblock In {\em CVPR}, 2019.

\bibitem{seff2017continual}
Ari Seff, Alex Beatson, Daniel Suo, and Han Liu.
\newblock Continual learning in generative adversarial nets.
\newblock {\em arXiv preprint arXiv:1705.08395}, 2017.

\bibitem{shi2022gradient}
Yuge Shi, Jeffrey Seely, Philip Torr, Siddharth N, Awni Hannun, Nicolas
  Usunier, and Gabriel Synnaeve.
\newblock Gradient matching for domain generalization.
\newblock In {\em ICLR}, 2022.

\bibitem{shin2017continual}
Hanul Shin, Jung~Kwon Lee, Jaehong Kim, and Jiwon Kim.
\newblock Continual learning with deep generative replay.
\newblock In {\em NeurIPS}, 2017.

\bibitem{thanh2020catastrophic}
Hoang Thanh-Tung and Truyen Tran.
\newblock Catastrophic forgetting and mode collapse in gans.
\newblock In {\em IJCNN}, 2020.

\bibitem{DFME}
Jean-Baptiste Truong, Pratyush Maini, Robert~J Walls, and Nicolas Papernot.
\newblock Data-free model extraction.
\newblock In {\em CVPR}, 2021.

\bibitem{yin2020dreaming}
Hongxu Yin, Pavlo Molchanov, Jose~M Alvarez, Zhizhong Li, Arun Mallya, Derek
  Hoiem, Niraj~K Jha, and Jan Kautz.
\newblock Dreaming to distill: Data-free knowledge transfer via deepinversion.
\newblock In {\em CVPR}, 2020.

\bibitem{KEGNET}
Jaemin Yoo, Minyong Cho, Taebum Kim, and U Kang.
\newblock Knowledge extraction with no observable data.
\newblock In {\em NeurIPS}, 2019.

\bibitem{WRN}
Sergey Zagoruyko and Nikos Komodakis.
\newblock Wide residual networks.
\newblock In {\em BMVC}, 2016.

\bibitem{zagoruyko2017paying}
Sergey Zagoruyko and Nikos Komodakis.
\newblock Paying more attention to attention: Improving the performance of
  convolutional neural networks via attention transfer.
\newblock In {\em ICLR}, 2017.

\bibitem{DDAD}
Haoran Zhao, Xin Sun, Junyu Dong, Milos Manic, Huiyu Zhou, and Hui Yu.
\newblock Dual discriminator adversarial distillation for data-free model
  compression.
\newblock {\em International Journal of Machine Learning and Cybernetics},
  2022.

\end{thebibliography}
}
\onecolumn
\ifarxiv \clearpage \appendix
\label{sec:appendix}
\section{Training Details:}
\begin{algorithm}[h]
\caption{Proposed DFKD method, with Memory-Buffer replay.}\label{alg:DFKD_MB}
\SetKwInput{KwData}{Input}
\SetKwInput{KwResult}{Output}
\KwData{$\mathcal{T}_{\theta_{\mathcal{T}}}$, $\mathcal{S}_{\theta_{\mathcal{S}}}$, $\mathcal{G}_{\theta_{\mathcal{G}}}$, $\mathcal{M}$, $\mathcal{E}_{max}$, $\mathcal{I}$,  $g$, $\alpha_{\mathcal{G}}$, $s$, $\alpha$, $\alpha_{\mathcal{S}}$, $f$}
\KwResult{$\mathcal{S}_{\theta_{\mathcal{S}}}$}
$\mathcal{E} = 1$\\
\While{$ \mathcal{E} \leq \mathcal{E}_{max}$}{\For{$\mathcal{I}$ iterations}{
    \For{$g$ iterations}{
    $z \sim \mathcal{N}(0,I)$ \\
    
    $\mathcal{L}_{\mathcal{G}} \leftarrow -\mathcal{D}( \mathcal{T}(\mathcal{G}_{\theta_{\mathcal{G}}}(z)), \mathcal{S}(\mathcal{G}_{\theta_{\mathcal{G}}}(z))) + \mathcal{L_{P}}(\mathcal{G}_{\theta_{\mathcal{G}}}(z))$ \\
    $\theta_{\mathcal{G}} \leftarrow \theta_{\mathcal{G}} - \alpha_{\mathcal{G}} \nabla_{\theta_{\mathcal{G}}}\mathcal{L}_{\mathcal{G}}$
    }
    \For{$s$ iterations}{$z \sim \mathcal{N}(0,I)$\\
    $\hat{x} \leftarrow \mathcal{G}_{\theta_{\mathcal{G}}}(z)$\\
    Compute $\mathcal{L}_{Acq}(\theta_{\mathcal{S}})$ using $\hat{x}$\\
    $\mathcal{L}_{\mathcal{S}} \leftarrow \mathcal{L}_{Acq}(\theta_{\mathcal{S}})$\\
    \If{$\mathcal{M}$ is not empty}{$\hat{x}_{m} \sim \mathcal{M}$ \\
    $\theta_{\mathcal{S}}^{\prime} \leftarrow \theta_{\mathcal{S}} - \alpha \nabla \mathcal{L}_{Acq}(\theta_{\mathcal{S}})$\\
    Compute $\mathcal{L}_{Ret}(\theta_{\mathcal{S}})$ and $\mathcal{L}_{Ret}(\theta_{\mathcal{S}}^{\prime})$ using $\hat{x}_{m}$\\
    $\mathcal{L}_{\mathcal{S}} \leftarrow \mathcal{L}_{\mathcal{S}} + \mathcal{L}_{Ret}(\theta_{\mathcal{S}}) + \mathcal{L}_{Ret}(\theta_{\mathcal{S}}^{\prime})$}
    $\theta_{\mathcal{S}} \leftarrow \theta_{\mathcal{S}} - \alpha_{\mathcal{S}}\nabla_{\theta_{\mathcal{S}}}\mathcal{L}_{S}$\\
       
    }
    }
    \If{$\mathcal{E}\mod f == 0$}{Update $\mathcal{M}$ with $x_{m}^{*}$, where, $ x_{m}^{*} \subseteq \hat{x}$}
    $\mathcal{E} \leftarrow \mathcal{E} + 1$}
\end{algorithm}
\begin{algorithm}[ht]
\caption{Proposed DFKD method, with Generative replay.}\label{alg:DFKD_GR}
\SetKwInput{KwData}{Input}
\SetKwInput{KwResult}{Output}
\KwData{$\mathcal{T}_{\theta_{\mathcal{T}}}$, $\mathcal{S}_{\theta_{\mathcal{S}}}$, $\mathcal{G}_{\theta_{\mathcal{G}}}$, $\mathcal{M}$, $\mathcal{E}_{max}$, $\mathcal{I}$,  $g$, $\alpha_{\mathcal{G}}$, $s$, $\alpha$, $\alpha_{\mathcal{S}}$, $f$, $s^{gp}_{max}$}
\KwResult{$\mathcal{S}_{\theta_{\mathcal{S}}}$}
$\mathcal{E} = 1$\\
\While{$ \mathcal{E} \leq \mathcal{E}_{max}$}{\For{$\mathcal{I}$ iterations}{
    \For{$g$ iterations}{
    $z \sim \mathcal{N}(0,I)$ \\
    
    $\mathcal{L}_{\mathcal{G}} \leftarrow -\mathcal{D}( \mathcal{T}(\mathcal{G}_{\theta_{\mathcal{G}}}(z)), \mathcal{S}(\mathcal{G}_{\theta_{\mathcal{G}}}(z))) + \mathcal{L_{P}}(\mathcal{G}_{\theta_{\mathcal{G}}}(z))$ \\
    $\theta_{\mathcal{G}} \leftarrow \theta_{\mathcal{G}} - \alpha_{\mathcal{G}} \nabla_{\theta_{\mathcal{G}}}\mathcal{L}_{\mathcal{G}}$
    }
    \For{$s$ iterations}{$z \sim \mathcal{N}(0,I)$\\
    $\hat{x} \leftarrow \mathcal{G}_{\theta_{\mathcal{G}}}(z)$\\
    Compute $\mathcal{L}_{Acq}(\theta_{\mathcal{S}})$ using $\hat{x}$\\
    $\mathcal{L}_{\mathcal{S}} \leftarrow \mathcal{L}_{Acq}(\theta_{\mathcal{S}})$\\
    $\hat{x}_{m} \sim \mathcal{M}$ \\
    $\theta_{\mathcal{S}}^{\prime} \leftarrow \theta_{\mathcal{S}} - \alpha \nabla \mathcal{L}_{Acq}(\theta_{\mathcal{S}})$\\
    Compute $\mathcal{L}_{Ret}(\theta_{\mathcal{S}})$ and $\mathcal{L}_{Ret}(\theta_{\mathcal{S}}^{\prime})$ using $\hat{x}_{m}$\\
    $\mathcal{L}_{\mathcal{S}} \leftarrow \mathcal{L}_{\mathcal{S}} + \mathcal{L}_{Ret}(\theta_{\mathcal{S}}) + \mathcal{L}_{Ret}(\theta_{\mathcal{S}}^{\prime})$\\
    $\theta_{\mathcal{S}} \leftarrow \theta_{\mathcal{S}} - \alpha_{\mathcal{S}}\nabla_{\theta_{\mathcal{S}}}\mathcal{L}_{S}$\\
    $s^{gp} = 0$\\
     \If{$\mathcal{E}\mod f == 0$ and $s^{gp} \leq s^{gp}_{max}$}{Train $\mathcal{M}$ with $\hat{x}_{m}$ and $x_{m}^{*}$, where, $ x_{m}^{*} \subseteq \hat{x}$\\
     $s^{gp} \leftarrow  s^{gp} +  1$\\}
    }
    }
    $\mathcal{E} \leftarrow \mathcal{E} + 1$}
\end{algorithm}
\subsection{Teacher Model Training Details}
We train the ResNet-34 \cite{he2016deep} teacher model for SVHN \cite{SVHN} and Tiny-ImageNet \cite{tiny}. For SVHN we use the  ResNet-34 model definition made available by Binci \textit{et al.}\footnote{\label{note1}\url{https://github.com/kuluhan/PRE-DFKD}} and for Tiny-ImageNet, we use the \texttt{torchvision} model definition from PyTorch\footnote{\url{https://pytorch.org/}}. To train the teacher models we use SGD optimizer with an initial learning rate of 0.1, momentum of 0.9 and a weight-decay of 5e-4, with a batch size of 128 for 400 epochs. Moreover, the learning rate is decayed at each iteration till 0, using cosine annealing.

\subsection{Student Model Training Details}
\label{sec:training_details}
For fair comparisons, we use the same Generator ($\mathcal{G}$) network (shown in Table \ref{tab:generator}) for all the methods. Unless not explicitly specified, for MB-DFKD \cite{binici2022preventing} and our method (w/ Memory Buffer), we maintain a memory buffer of size 10 and update the memory buffer at a frequency of $f=5$, following previous work \cite{binici2022preventing} (Algorithm \ref{alg:DFKD_MB}). Also, for PRE-DFKD \cite{binici2022robust} and our method (w/ Generative Replay), we use the same VAE architecture (as in Table \ref{tab:generator} (Decoder) and \ref{tab:vae_encoder} (Encoder)), from \cite{binici2022robust}, to transfer the pseudo samples as memory, and use the decoder part (same as the generator architecture in Table \ref{tab:generator}) to replay the learnt distribution, with the VAE update parameters of $f=1$ and $s_{max}^{gp} = 4$ (Algorithm \ref{alg:DFKD_GR}), following previous works \cite{binici2022robust}. 
For all the methods and datasets, we use SGD optimizer with a momentum of 0.9 and a variable learning rate ($\alpha_{\mathcal{S}}$) with cosine annealing starting from 1e-1 and annealing it at each epoch to 0 to optimize the student parameters ($\theta_{\mathcal{S}}$). For the one-step gradient descent, we use a learning rate ($\alpha$) of 0.9. Furthermore, we use Adam optimizer with a learning rate ($\alpha_{\mathcal{G}}$) of 0.02 to optimize the Generator ($\mathcal{G}$). We test all our methods primarily on SVHN \cite{SVHN}, CIFAR10 \cite{CIFAR}, CIFAR100 \cite{CIFAR}, and Tiny-ImageNet \cite{tiny} for 200, 200, 400, and 500 epochs ($\mathcal{E}_{max}$), respectively. Our experiments were run on a mixture of Nvidia RTX2080Ti (11GB) and RTX3090 (24GB) GPUs. However, all our experiments incured not more than 11.5 GB of VRAM.
\begin{table}[ht]
\centering
\caption{Generator Network ($\mathcal{G}$) and Generative Replay (VAE \cite{VAE}) Decoder Architecture.}
\label{tab:generator}
\resizebox{0.65\textwidth}{!}{%
\begin{tabular}{@{}ll@{}}
\toprule
\textbf{Output Size}                                 & \textbf{Layers}                                                                                \\ \midrule
$1000$                                      & Noise ($z \sim \mathcal{N}(0,I)$)                                                     \\
$128 \times  h / 4 \times  w / 4$ & \textit{Linear}, \textit{BatchNorm1D}, \textit{Reshape}                               \\
$128 \times   h / 4 \times w / 4$ & \textit{SpectralNorm (Conv ($3 \times 3$))}, \textit{BatchNorm2D}, \textit{LeakyReLU} \\
$128 \times  h / 2 \times w / 2$  & \textit{UpSample} ($2 \times$)                                                        \\
$64 \times  h / 2 \times w / 2$   & \textit{SpectralNorm (Conv ($3 \times 3$))}, \textit{BatchNorm2D}, \textit{LeakyReLU} \\
$64 \times  h  \times w $         & \textit{UpSample} ($2 \times$)                                                        \\
$3 \times  h  \times w $          & \textit{SpectralNorm (Conv ($3 \times 3$))}, \textit{TanH}, \textit{BatchNorm2D}      \\ \bottomrule
\end{tabular}%
}
\end{table}

\begin{table}[ht]
\centering
\caption{Generative Replay (VAE \cite{VAE}) Encoder Architecture.}
\label{tab:vae_encoder}
\resizebox{0.65\textwidth}{!}{%
\begin{tabular}{@{}ll@{}}
\toprule
\textbf{Output Size}        & \textbf{Layers}                                                                       \\ \midrule
$3 \times h \times w$       & Input Example                                                                         \\
$64 \times h \times w$      & \textit{SpectralNorm(Conv ($3 \times 3$))}, \textit{BatchNorm2D}, \textit{LeakyReLU} \\
$128 \times h \times w$     & \textit{SpectralNorm(Conv ($3 \times 3$))}, \textit{BatchNorm2D}, \textit{LeakyReLU} \\
$128 \times h/2 \times w/2$ & \textit{DownSample} ($0.5 \times$)                                                             \\
$128 \times h/2 \times w/2$ & \textit{SpectralNorm(Conv ($3 \times 3$))}, \textit{BatchNorm2D}\\
$128 \times h/4 \times w/4$ & \textit{DownSample} ($0.5 \times$)                                                             \\
$\{1000, 1000\}$            & \textit{Reshape}, \textit{Linear}                                                     \\ \bottomrule
\end{tabular}%
}
\end{table}

\section{Attribution of Existing Assets:}
\subsection{Code-Base:}
The code-base used to experiment with proposed method is adapted from the GitHub\textsuperscript{\ref{note1}} repository of Binci \textit{et al.} \cite{binici2022robust}. 
\subsection{Pre Trained Teacher Model}
The CIFAR10 pretrained \cite{CIFAR} Teacher models of ResNet-34 and WRN-40-2 \cite{WRN} are used used from the GitHub\footnote{\url{https://github.com/zju-vipa/CMI}} repository made available by Fang \textit{et al.} \cite{fang2021contrastive}. For the ResNet-34 Teacher model, pretrained on CIFAR100 \cite{CIFAR}, we used the model made available by Binci \textit{et al.}\textsuperscript{\ref{note1}} \cite{binici2022robust}.

\section{Extended Results}
\paragraph{}
In Figure \ref{fig:wrn_cumulative}, we visualize the Cumulative Mean Accuracies (\%) across the training epochs with Buffer-based and Generative Replay. The plots in Figure \ref{fig:wrn_cumulative} complement the ones shown in Figure \ref{fig:wrn_cifar10} of the main manuscript.
\paragraph{}
Based on the similarity of the Tiny-ImageNet teacher accuracy ($\mathcal{T}_{Acc})$ of the methods proposed and reported by Li \textit{et al.} \cite{CuDFKD}, we compare our methods with the accuracies reported by them.
\begin{table}[ht]
\centering
\label{tab:tiny_extended}
\resizebox{0.7\textwidth}{!}{%
\begin{tabular}{lcc}
\toprule
\textbf{Method}                                 & {\color[HTML]{656565} \textbf{Teacher Accuracy (\%) ($\mathcal{T}_{Acc}$)}} & \textbf{Student Accuracy (\%) ($\mathcal{S}_{Acc}$)} \\ \midrule
ADI\textsuperscript{a} \cite{yin2020dreaming}                                              & {\color[HTML]{656565} 61.47}                                                & 6.00                                                 \\
CMI\textsuperscript{a} \cite{fang2021contrastive}                                              & {\color[HTML]{656565} 61.47}                                                & 1.85                                                 \\
 \midrule
DAFL\textsuperscript{b} \cite{DAFL}                                             & {\color[HTML]{656565} 60.83}                                                & 35.46                                                \\
DFAD\textsuperscript{b} \cite{fang2019data}                                             & {\color[HTML]{656565} 60.83}                                                & 19.60                                                \\
DFQ\textsuperscript{a} \cite{choi2020data}                                             & {\color[HTML]{656565} 61.47}                                                & 41.30                                                \\
CuDFKD\textsuperscript{a} \cite{CuDFKD}                                          & {\color[HTML]{656565} 61.47}                                                & 43.42                                                \\ \midrule
\rowcolor[HTML]{EFEFEF} 
\textit{\textbf{Ours-1 (w/ Memory Bank)}}       & {\color[HTML]{656565} 60.83}                                                & 47.96                                                \\
\rowcolor[HTML]{EFEFEF} 
\textit{\textbf{Ours-2 (w/ Generative Replay)}} & {\color[HTML]{656565} 60.83}                                                & 49.88                                                \\ \bottomrule
\end{tabular}%
}
\caption{Classification accuracy (in \%) of the student trained using various DFKD methods on Tiny-ImageNet \cite{tiny} with ResNet-34 \cite{he2016deep} as the teacher and ResNet-18 \cite{he2016deep} as the student. \textsuperscript{a} and \textsuperscript{b} denote results obtained from \cite{CuDFKD} and our implementation, respectively.}
\end{table}

\begin{figure}[ht]
    \centering
    \includegraphics[width=\textwidth]{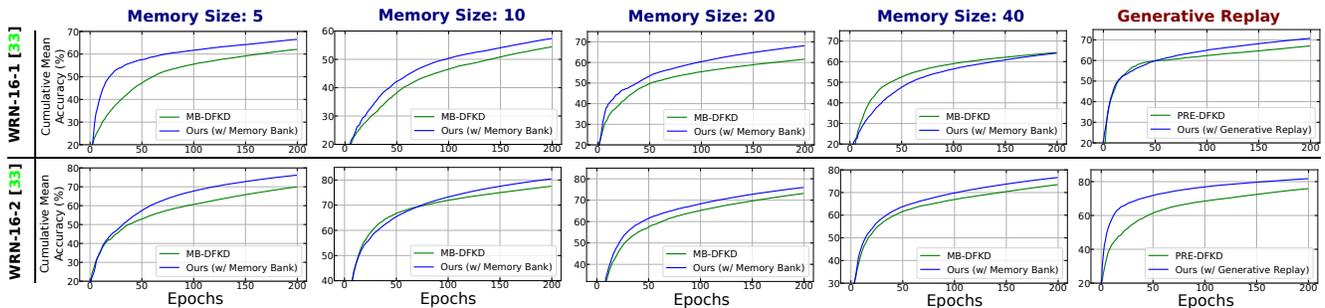}
    \caption{Cumulative Mean Accuracy (\%) evolution of Wide-ResNet (WRN) \cite{WRN}. The WRN-16-1 (\textit{top-row}), and WRN-16-2 (\textit{bottom-row}) networks are distilled by a WRN-40-2 teacher network pre-trained on CIFAR10 ($\mathcal{T}_{Acc} = 94.87\%$). Each column represent the learning curves with the Buffer-based (with different memory buffer sizes) and Generative replay schemes. The proposed method is in \textbf{\color[HTML]{3b80ee} Blue}.}
    \label{fig:wrn_cumulative}
\end{figure}
\newpage
\begin{customlemma}{1}
\label{lemma:taylor_series_appn}
If $\mathcal{L}_{Ret}$ has Lipschitz Hessian, \textit{i.e.}, $\lVert \nabla^{2}\mathcal{L}_{Ret}(\theta_{1}) - \nabla^{2}\mathcal{L}_{Ret}(\theta_{2}) \rVert \leq \rho \lVert \theta_{1} - \theta_{2} \rVert$ for some $\rho > 0$, then:
\begin{align*}
    \nabla\mathcal{L}_{Ret}(\theta + \mathbf{\phi}_{\theta}) = \nabla\mathcal{L}_{Ret}(\theta) +  \nabla^{2}\mathcal{L}_{Ret}(\theta)\mathbf{\phi}_{\theta} + \mathcal{O}(\lVert \mathbf{\phi}_{\theta} \rVert^{2}).
\end{align*}
For instance, when $\phi_{\theta} = -\alpha \nabla \mathcal{L}_{Acq}(\theta)$, we have,
\begin{align*}
    \nabla\mathcal{L}_{Ret}(\theta -\alpha \nabla \mathcal{L}_{Acq}(\theta)) = & \nabla\mathcal{L}_{Ret}(\theta) -  \alpha \nabla^{2}\mathcal{L}_{Ret}(\theta) \nabla \mathcal{L}_{Acq}(\theta) + \mathcal{O}(\alpha^{2}). 
\end{align*}
\end{customlemma}
\begin{proof}
    Applying the fundamental theorem of calculus to each component of $\mathcal{L}_{Ret}$, we have:  

    \begin{align}
        \nabla \mathcal{L}_{Ret}(\theta + \phi_{\theta}) =  \nabla \mathcal{L}_{Ret}(\theta) + \nabla^{2}\mathcal{L}_{Ret}(\theta)\phi_{\theta} + \int_{k=0}^{1}( \nabla^{2} \mathcal{L}_{Ret}(\theta + k\phi_{\theta}) - \nabla^{2}\mathcal{L}_{Ret}(\theta))\phi_{\theta}dk.
    \end{align}
Omitting the subscript $Ret$ for brevity,
    \begin{align}
        \implies   \lVert \nabla \mathcal{L}(\theta + \phi_{\theta}) - (\nabla \mathcal{L}(\theta) + \nabla^{2}\mathcal{L}(\theta)\phi_{\theta}) \rVert &= \lVert \int_{k=0}^{1}( \nabla^{2} \mathcal{L}(\theta + k\phi_{\theta}) - \nabla^{2}\mathcal{L}(\theta))\phi_{\theta}dk \rVert \\
       \implies   \lVert \nabla \mathcal{L}(\theta + \phi_{\theta}) - (\nabla \mathcal{L}(\theta) + \nabla^{2}\mathcal{L}(\theta)\phi_{\theta}) \rVert & \leq \int_{k=0}^{1} \lVert ( \nabla^{2} \mathcal{L}(\theta + k\phi_{\theta}) - \nabla^{2}\mathcal{L}(\theta))\phi_{\theta} \rVert dk\\
        \implies   \lVert \nabla \mathcal{L}(\theta + \phi_{\theta}) - (\nabla \mathcal{L}(\theta) + \nabla^{2}\mathcal{L}(\theta)\phi_{\theta}) \rVert & \leq \int_{k=0}^{1} \rho \lVert k \phi_{\theta} \rVert. \lVert \phi_{\theta} \rVert dk \qquad \text{from $\rho$-Lipschitzness} \\
        \implies   \lVert \nabla \mathcal{L}(\theta + \phi_{\theta}) - (\nabla \mathcal{L}(\theta) & + \nabla^{2}\mathcal{L}(\theta)\phi_{\theta}) \rVert \leq \frac{\rho}{2} \lVert \phi_{\theta} \rVert^{2}.
   \end{align}
\end{proof}
\begin{customthm}{1}
\label{lemma:theorem1_proof}
If $\theta^\prime = \theta - \alpha \nabla \mathcal{L}_{Acq}(\theta)$, denotes the one step gradient descent on $\theta$ with the objective $\mathcal{L}_{Acq}(\theta)$, where $\alpha$ is a scalar, and $\nabla\mathcal{L}_{Acq}(\theta)$ denotes the gradients of $\mathcal{L}_{Acq}$ at $\theta$, then: 
\begin{align*}
        \frac{\partial \mathcal{L}_{Ret}(\theta^{\prime})}{\partial \theta} = \nabla \mathcal{L}_{Ret}(\theta) - \alpha \nabla^{2} \mathcal{L}_{Ret}(\theta).\nabla \mathcal{L}_{Acq}(\theta) - \alpha \nabla^{2} \mathcal{L}_{Acq}(\theta).\nabla \mathcal{L}_{Ret}(\theta) + \mathcal{O}(\alpha^{2}).
\end{align*}
\end{customthm}
\begin{proof}
We have
\begin{align}
    \frac{\partial \mathcal{L}_{Ret}(\theta^{\prime})}{\partial \theta} &= \nabla \mathcal{L}_{Ret}(\theta^{\prime}). \frac{\partial \theta^{\prime}}{\partial \theta} \\
    \implies \frac{\partial \mathcal{L}_{Ret}(\theta^{\prime})}{\partial \theta} &= \nabla \mathcal{L}_{Ret}(\theta^{\prime}). \frac{\partial( \theta - \alpha \nabla \mathcal{L}_{Acq}(\theta))}{\partial \theta} \\
    \implies \frac{\partial \mathcal{L}_{Ret}(\theta^{\prime})}{\partial \theta} &= \nabla \mathcal{L}_{Ret}(\theta^{\prime}).(I - \alpha \nabla^{2}\mathcal{L}_{Acq}(\theta)) \label{eq:partial}
\end{align}

Using Lemma \ref{lemma:taylor_series_appn}, we substitute the value of $\nabla \mathcal{L}_{Ret}(\theta^{\prime})$, where $\theta^\prime = \theta - \alpha \nabla \mathcal{L}_{Acq}(\theta)$ in (\ref{eq:partial}), and obtain:
\begin{align}
    \frac{\partial \mathcal{L}_{Ret}(\theta^{\prime})}{\partial \theta}  &= \overbrace{(\nabla \mathcal{L}_{Ret}(\theta) + \nabla^{2}\mathcal{L}_{Ret}(\theta).\underbrace{(\theta^{\prime} - \theta)}_{= -\alpha \nabla \mathcal{L}_{Acq}(\theta)} + \underbrace{\mathcal{O}(\lVert \theta^{\prime} - \theta\rVert^{2})}_{=\mathcal{O}(\alpha^2)})}^{=\nabla\mathcal{L}_{Ret}(\theta^{\prime})}.(I - \alpha \nabla^{2}\mathcal{L}_{Acq}(\theta))\\
    \implies \frac{\partial \mathcal{L}_{Ret}(\theta^{\prime})}{\partial \theta}  &=  \nabla\mathcal{L}_{Ret}(\theta) + \nabla^{2}\mathcal{L}_{Ret}(\theta).\underbrace{(\theta^{\prime} - \theta)}_{= -\alpha \nabla \mathcal{L}_{Acq}(\theta)} - \alpha \nabla^{2}\mathcal{L}_{Acq}(\theta)\nabla \mathcal{L}_{Ret}(\theta) + \mathcal{O}(\alpha^{2}) \\
     \implies \frac{\partial \mathcal{L}_{Ret}(\theta^{\prime})}{\partial \theta}  &=  \nabla\mathcal{L}_{Ret}(\theta) - \alpha \nabla^{2}\mathcal{L}_{Ret}(\theta)\nabla \mathcal{L}_{Acq}(\theta)- \alpha \nabla^{2}\mathcal{L}_{Acq}(\theta)\nabla \mathcal{L}_{Ret}(\theta) + \mathcal{O}(\alpha^{2}) \\ 
     \implies \frac{\partial \mathcal{L}_{Ret}(\theta^{\prime})}{\partial \theta} &=  \nabla\mathcal{L}_{Ret}(\theta) - \alpha \underbrace{(\overbrace{\nabla^{2}\mathcal{L}_{Ret}(\theta)\nabla \mathcal{L}_{Acq}}^{Hessian \ Product-1} - \overbrace{\nabla^{2}\mathcal{L}_{Acq}\nabla \mathcal{L}_{Ret}(\theta)}^{Hessian \ Product-2})}_{Gradient \ Matching} + \mathcal{O}.(\alpha^{2}) 
     \label{eq:lret}
\end{align}

Note that, Lemma \ref{lemma:taylor_series_appn} provides an efficient way to obtain $Hessian \ Product-1$ (highlighted in (\ref{eq:lret})) by computing the gradient of $\mathcal{L}_{Ret}$ at $\theta^{\prime}$, thus, eradicating the time and memory overhead of explicitly computing $Hessian \ Product-1$. Hence, we have:
\begin{align}
    \frac{\partial \mathcal{L}_{Ret}(\theta^{\prime})}{\partial \theta} =  \nabla \mathcal{L}_{Ret}(\theta) - \alpha \nabla^{2}\mathcal{L}_{Ret}(\theta)\nabla \mathcal{L}_{Acq}(\theta)  - \alpha \nabla^{2}\mathcal{L}_{Acq}\nabla \mathcal{L}_{Ret}(\theta) + \mathcal{O}(\alpha^{2}).
\end{align}
\end{proof}

 \fi

\end{document}